\documentclass[11pt,a4paper]{article}
\usepackage[hyperref]{acl2020}
\usepackage{times}
\usepackage{latexsym}

\usepackage{microtype}

\usepackage[utf8]{inputenc} 

\usepackage{array}
\usepackage{amssymb}
\usepackage{amsmath}
\usepackage{diagbox}
\usepackage{graphicx}
\usepackage{multirow}
\usepackage{cjhebrew}
\usepackage{soul}
\usepackage{enumitem}
\usepackage{multirow}
\usepackage{subcaption}
\usepackage{booktabs}
\usepackage{IEEEtrantools}
\usepackage{stackengine}
\usepackage[title]{appendix}
\usepackage{longtable}

\renewcommand\cite{\citep}                   % (Author, Year)
\newcommand{\fone}{F$_1$}                    % F1 score
\newcommand{\bert}{mBERT}
\newcommand{\nlm}{LM}
\newcommand{\plaincomma}{{\normalfont, }}
\newcolumntype{R}[1]{>{\raggedleft\arraybackslash}p{#1}}  % right-aligned col
\newcolumntype{C}[1]{>{\centering\arraybackslash}p{#1}}  % center-aligned col
\definecolor{lemonchiffon}{HTML}{FFFACD}
\definecolor{whitesmoke}{HTML}{F5f5f5}
\definecolor{dodgerblue}{HTML}{1E90FF}
\definecolor{tomato}{HTML}{FF6347}
\definecolor{mediumseagreen}{HTML}{3CB371}
\definecolor{pink}{HTML}{FFC0CB}

\sethlcolor{pink}

\aclfinalcopy

\title{A multilabel approach to morphosyntactic probing}

\author{
  Naomi Tachikawa Shapiro\plaincomma Amandalynne Paullada\plaincomma\and Shane Steinert-Threlkeld \\
  Department of Linguistics, University of Washington, Seattle, WA, USA \\
  \texttt{\{tsnaomi,paullada,shanest\}@uw.edu}
  }

\begin{document}

\maketitle

\begin{abstract}
  We introduce a multilabel probing task to assess the morphosyntactic representations of word embeddings from multilingual language models. We demonstrate this task with multilingual BERT \cite{Devlin2018}, training probes for seven typologically diverse languages of varying morphological complexity: Afrikaans, Croatian, Finnish, Hebrew, Korean, Spanish, and Turkish. Through this simple but robust paradigm, we show that multilingual BERT renders many morphosyntactic features easily and simultaneously extractable (e.g., gender, grammatical case, pronominal type). We further evaluate the probes on six ``held-out'' languages in a zero-shot transfer setting: Arabic, Chinese, Marathi, Slovenian, Tagalog, and Yoruba. This style of probing has the added benefit of revealing the linguistic properties that language models recognize as being shared across languages. For instance, the probes performed well on recognizing nouns in the held-out languages, suggesting that multilingual BERT has a conception of \emph{noun}-hood that transcends individual languages; yet, the same was not true of adjectives.
  \end{abstract}

\section{Introduction} \label{sec:intro}

  Morphologically rich languages present unique challenges to natural language processing. These languages typically exhibit complex agreement patterns and their high diversity of inflected forms can lead to sparse examples of vocabulary words in training data, even in large corpora \citep{Blevins2019,Gerz2018}. It is therefore worthwhile to explore how neural language models (\nlm s), which serve as the foundation of many state-of-the-art systems, handle the morphological complexity of diverse languages.
  
  Morphosyntactic features of natural languages bear meaningful information that is useful for downstream tasks, such as machine translation, question answering, and language generation. Research has shown that adding morphological supervision through multi-task training regimes \citep{Blevins2019} or morphologically-informed tokenization \citep{Klein2020,Park2020} can improve the quality of multilingual language models. Nonetheless, recent work has shown that \nlm s trained without explicit morphological supervision can still produce useful representations that capture morphosyntactic phenomena.\cite[e.g.,][]{Bacon2019,Pires2019,Dufter2020}.

  To further these investigations, we have developed a multilabel probing task to assess the morphosyntactic representations of contextualized word embeddings at a wide scale. This work is premised on the intuition that, if a simple model (a ``probe'') can easily extract linguistic properties from embeddings, this indicates that the \nlm\ has learned to encode those features in some fashion \cite[cf.][]{Conneau2018a,Hupkes2018,Liu2019a}. In particular, we show how a simple multilabel paradigm can shed light on the morphosyntactic representations of language models, both holistically and at the level of individual features. Our contributions are threefold: 

  \begin{itemize}
    \item We introduce an efficient and robust probing paradigm for extracting multiple morphosyntactic features, which we demonstrate with multilingual BERT \citep{Devlin2018} and seven typologically diverse languages: Afrikaans, Croatian, Finnish, Hebrew, Korean, Spanish, and Turkish.

    \item We evaluate the probes on six ``held-out'' languages---Arabic, Chinese, Marathi, Slovenian, Tagalog, and Yoruba---showing how this paradigm can additionally be used to illuminate the properties that \bert\ represents similarly cross-linguistically.
    
    \item We release our code and plan to release out multilabel probe predictions to encourage more in-depth analyses of multilingual BERT and to guide future probing efforts.\footnote{\url{https://github.com/tsnaomi/morph-bert}}
  \end{itemize}

  Throughout the paper, we also tackle questions of memorization and probe complexity by evaluating the probes on control tasks \cite[cf.][]{Hewitt2019b}. The remainder of this paper is structured as follows: Section \ref{sec:related-work} reviews related work, motivating \S\ref{sec:tagging}, which introduces the multilabel morphosyntactic probing task. Section \ref{sec:setup} then outlines the data and multilabel models that we use to probe multilingual BERT. In \S\ref{sec:mono}, we demonstrate our probing paradigm in a set of \emph{monolingual} experiments, training and evaluating separate probes for each of the seven languages. In \S\ref{sec:multi}, we delve into whether \emph{multilingual} probes yield comparable insights. Then, in a set of \emph{crosslingual} experiments, \S\ref{sec:transfer} evaluates how the monolingual and multilingual probes handle the six held-out languages. Finally, \S\ref{sec:discussion} discusses our findings while \S\ref{sec:conclusion} concludes the paper.

\section{Related work} \label{sec:related-work}
    
    % Morphological tagging
      Numerous studies in recent years have sought to study the linguistic properties captured by neural language models \cite[e.g.,][]{Conneau2018a,Gulordava2018,Hupkes2018,Marvin2018,Zhang2018,Bacon2019,Futrell2019,Hewitt2019a,Jawahar2019,Liu2019a,Tenney2019,Chi2020}. 
          
      In the morphology domain, the LINSPECTOR suite by \citet{Sahin2020} probes 24 languages via 15 linguistic tasks, including multiple tasks to identify morphological features. In a similar vein, \citet{Edmiston2020} uses several morphological prediction tasks to inspect embeddings from five monolingual Transformer-based language models, focusing exclusively on Indo-European languages. The probing paradigm proposed in this paper builds on the work of \citet{Sahin2020} and \citet{Edmiston2020}, but consolidates morphosyntactic feature prediction under a single task.

    \subsection{Vying for control}

      Recent work has sought to curtail how much probes memorize about linguistic tasks to ensure that they \emph{reflect} information available in their input embeddings. In other words, probes should be extractive rather than learn\`ed themselves. Efforts to minimize memorization have included reducing the training data to probes \cite{Zhang2018} and limiting probe complexity, such as through dropout \cite[e.g.,][]{Belinkov2017a,Belinkov2017b,Sahin2020} and the use of simpler architectures (e.g., a linear layer instead of a multilayer perceptron, as in \citealt{Alain2018} and \citealt{Liu2019a}).

      To guide the design and interpretation of probes, \citet{Hewitt2019b} propose supplementing diagnostic tasks with \emph{control tasks}, where a probe is trained to predict random outputs within the same output space as the diagnostic task, given the same embeddings. If the probe performs well on the control task, they caution that it has the capacity to memorize the linguistic features under consideration; conversely, if the probe does well on the diagnostic task but poorly on the control task, then it is a reliable diagnostic of linguistic representations in the embeddings (though see \citealt{Pimentel2020b,Pimentel2020a} for interesting discussions). \citeauthor{Hewitt2019b} operationalize this comparison as \emph{selectivity}, the difference between the control task and diagnostic task performance. The greater the selectivity, the more a probe is said to ``express'' the information encoded in its inputs. In this paper, we design a control task to complement multilabel probing.

\begin{figure*}[t] % vectors
  \centering
  \includegraphics[width=14cm]{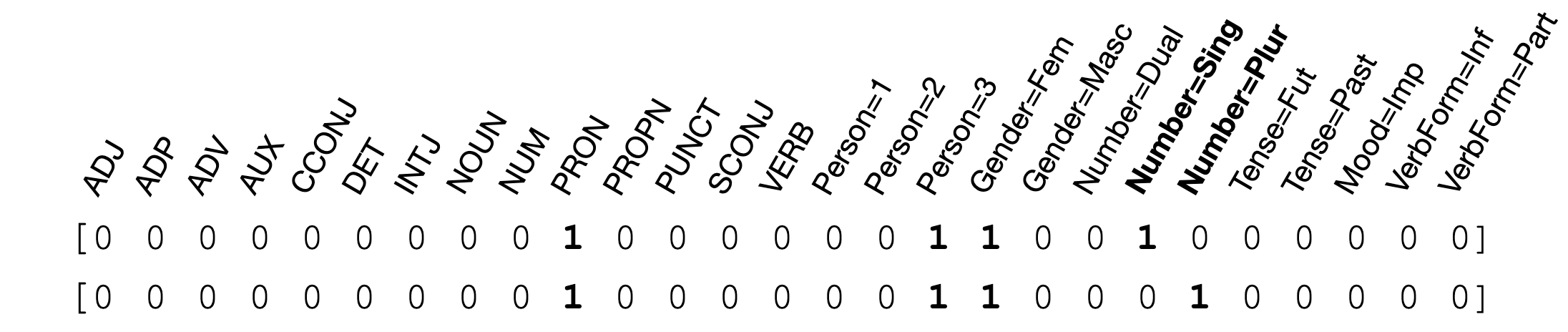}
  \caption{
  Hypothetical multi-hot encoded vectors for the Hebrew \textit{3.sing.fem} pronoun \<hy'> \emph{hi} (top) and \textit{3.plur.fem} pronoun \<hN> \emph{hen} (bottom). The two vectors differ only with respect to the two cells indicating ``singular'' and ``plural'', reflecting their otherwise similar grammatical characteristics.}
  \label{fig:vectors} 
\end{figure*}

\section{Multilabel morphosyntactic probing} \label{sec:tagging}

  We propose using multilabel morphosyntactic tagging as a diagnostic task to assess the morphosyntactic representations of neural language models. To do so, we hold contextualized word embeddings constant, then train linear classifiers on top of them \cite[cf.][]{Liu2019a,Hupkes2018} to perform the morphosyntactic tagging task. In its objective, morphosyntactic tagging resembles the second SIGMORPHON 2019 shared task, which called for labeling words in a sentence with their morphosyntactic descriptions \cite{McCarthy2019}.

  It is easy to imagine doing morphosyntactic tagging in a traditional multiclass fashion, where we train separate probes to identify different features, such as part of speech, gender, or number \cite[cf.][]{Sahin2020,Edmiston2020}. Alternatively, we could train a single probe to extract complex labels like \texttt{def.sing.masc.noun} and \texttt{3rd.plur.masc.past.verb}. Thus, each word would have a single correct label and a final softmax layer would output the probability of each class being the correct one. However, a drawback to this approach is that, depending on the number of properties we would like to identify, this can result in a combinatoric nightmare \includegraphics[height=1em]{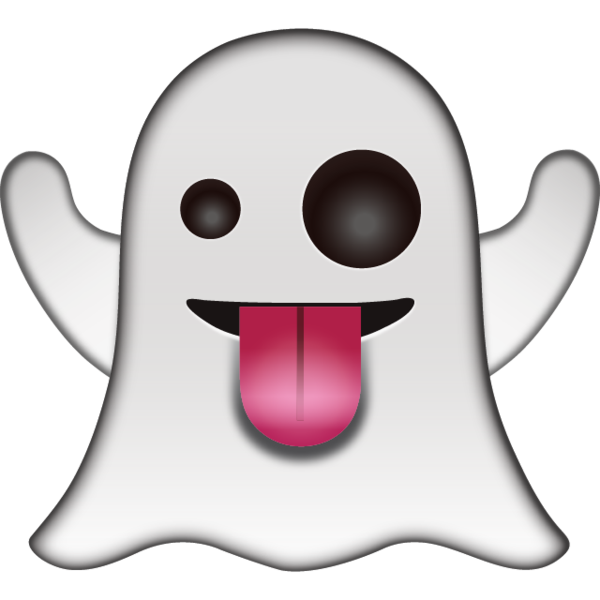}, with few training examples per class.
  
  To overcome this limitation, we frame morphosyntactic tagging as a word-level multilabel task, allowing for a token to receive multiple feature labels (e.g., both \texttt{Person=1} and \texttt{Number=Sing}). Target vectors are multi-hot encoded, such that a cell is 1 if a specific feature label applies to a word and 0 otherwise. A multilabel paradigm allows us to encode features with multiple or ambiguous values (e.g., \texttt{Gender=Fem,Masc}; a.k.a. multi-valued features) and enables a close inspection of learnt agreement and feature co-occurrence patterns. Figure \ref{fig:vectors} illustrates hypothetical gold vectors for two Hebrew pronouns that differ only in number.

  \subsection{Notation and nomenclature}

\begin{figure}[h] % feature label
      \centering
      \includegraphics[width=3.5cm]{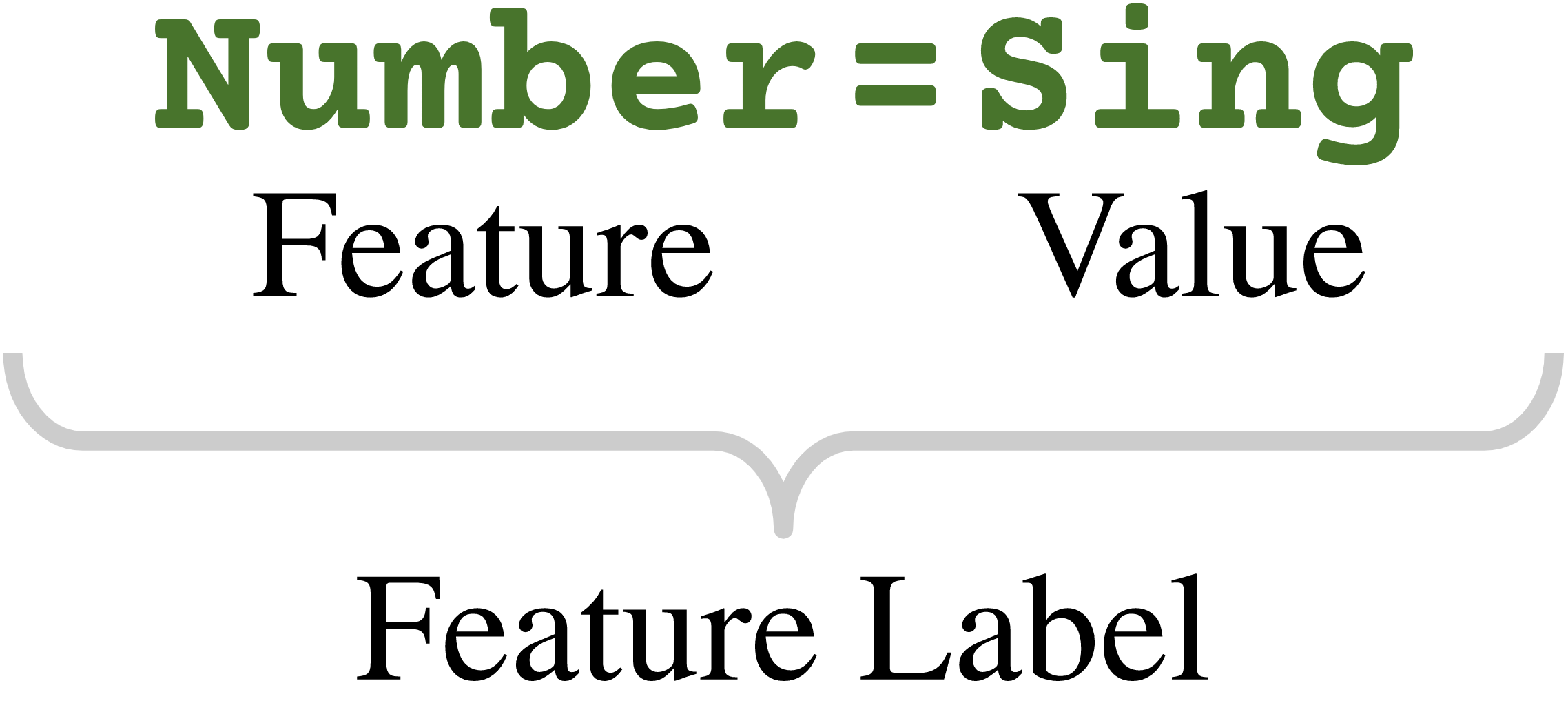}
      \caption{Anatomy of a feature label.}
      \label{fig:feature-label}
    \end{figure}

    We define a feature label as the conjunction of a linguistic feature  (e.g., \emph{number}) and a possible realized value of that feature (e.g., \emph{singular}), as depicted in Figure \ref{fig:feature-label}. Multiple feature labels can correspond to the same feature (e.g., \texttt{Number=Sing} and \texttt{Number=Plur}).\footnote{We drop \texttt{POS=} from part-of-speech labels, conforming to UPOS notation (e.g., \texttt{NOUN} instead of \texttt{POS=NOUN}).} We define $F$ as the set of feature labels $\{f_1, \ldots, f_{|F|}\}$ that we use to identify morphosyntactic properties from contextualized word embeddings.

    % \break
    Assuming a vocabulary of word types $V$, let $\mathbf{s} = s_1\ldots s_{|\mathbf{s}|}$ denote a specific sentence and $r_i$ denote the contextualized representation of each token $s_i$, such that $s_i\in V$. The inputs to the probe are therefore the embeddings $r_i \in\mathbb{R}^d$. In the multilabel morphosyntactic tagging task, we define the target output of each embedding $r_i$ as a multi-hot encoded vector $\mathbf{y}^i = y^i_1\ldots y^i_{|F|}$, where $F$ is the aforementioned set of feature labels. We encode $y^i_j$ as 1 if the feature label $f_j \in F$ describes the token $s_i$ and 0 otherwise.

  \subsection{Multilabel evaluation}
 
    The multilabel paradigm lends itself well to analyzing features both holistically and at a granular level. We can analyze individual features by calculating precision, recall, and \fone\ for each feature label $f$ separately. 
    Furthermore, we can glean the overall or \emph{micro-averaged} performance of a probe by first tallying the true positives (TP), false positives (FP), and false negatives (FN) across the features:
    \begin{align}
      \label{eq:micro}
      \begin{split}
        \text{F}_{1micro} &= 2 \times \frac{\text{P}_{micro} \times \text{R}_{micro}}{\text{P}_{micro} + \text{R}_{micro}}\quad\text{where} \\ \\
        \text{P}_{micro} &= \frac{\sum_f TP_f}{\sum_f TP_f + FP_f} \\ \\
        \text{R}_{micro} &= \frac{\sum_f TP_f}{\sum_f TP_f + FN_f}
      \end{split}
    \end{align}

    It is also possible to take the \emph{macro-average} over feature label scores (e.g., averaging the \fone\ scores across the labels). For simplicity, we focus on micro-averages in this paper. % Micro-averages are more impervious to one-off failures to extract a specific feature label

\begin{table*}[t] % seen
  {
    \centering
    \begin{tabular}{llc*{6}{r}} 
      \toprule
      \multicolumn{1}{c}{\multirow{2}{*}{Language}}   & \multirow{2}{*}{Genus} & \multirow{2}{*}{$|F|$}    & \multicolumn{2}{c}{Train}                                & \multicolumn{2}{c}{Dev}                                 & \multicolumn{2}{c}{Test}               \\ 
      \cmidrule(rl){4-5} \cmidrule(rl){6-7} \cmidrule(rl){8-9}
      \multicolumn{1}{c}{}                            &                        &                           & \multicolumn{1}{r}{Sentences} & \multicolumn{1}{r}{Tokens}   & \multicolumn{1}{r}{Sentences} & \multicolumn{1}{r}{Tokens}   & \multicolumn{1}{r}{Sentences} & \multicolumn{1}{r}{Tokens} \\ 
      \midrule
      Afrikaans                                       & Germanic               & 53                        & 800                        & 21,160                      & 194                        & 5,317                       & 425                        & 10,065   \\
      Croatian                                        & Slavic                 & 66                        & 800                        & 17,811                      & 960                        & 22,292                      & 1,136                      & 24,260   \\
      Finnish                                         & Finnic                 & 89                        & 800                        & 10,786                      & 1,363                      & 18,311                      & 1,553                      & 21,069   \\
      Hebrew                                          & Semitic                & 53                        & 800                        & 16,061                      & 484                        & 8,358                       & 491                        & 8,829    \\
      Korean                                          & Korean                 & 35                        & 800                        & 13,177                      & 100                        & 1,679                       & 100                        & 1,728    \\
      Spanish                                         & Romance                & 63                        & 800                        & 24,345                      & 1,654                      & 52,161                      & 1,719                      & 52,429   \\
      Turkish                                         & Turkic                 & 64                        & 800                        & 8,244                       & 983                        & 9,768                       & 981                        & 9,794    \\ 
      Multilingual                                    & n/a                    & 72                        & 4,800                      & 98,297                      & 5,638                      & 116,207                     & n/a                        & n/a      \\
      \bottomrule
    \end{tabular}  
  }
  \caption{Composition of the training and evaluation data for the monolingual and multilingual probes.}
  \label{tab:data-seen}
\end{table*}

\begin{table}[t] % held out
  {
    \centering
    \begin{tabular}{llrr} 
      \toprule
      \multicolumn{1}{c}{\multirow{2}{*}{Language}}   & \multirow{2}{*}{Genus} & \multicolumn{2}{c}{Test}             \\ 
      \cmidrule(rl){3-4}
      \multicolumn{1}{c}{}                            &                        & \multicolumn{1}{r}{Sentences} & \multicolumn{1}{r}{Tokens} \\ 
      \midrule
      Arabic                                          & Semitic                & 675                        & 24,195  \\
      Chinese                                         & Chinese                & 1,000                      & 21,415  \\
      Korean                                          & Korean                 & 1,000                      & 16,584  \\
      Marathi                                         & Indic                  & 47                         & 376     \\
      Slovenian                                       & Slavic                 & 995                        & 9,880   \\
      Tagalog                                         & GCP                    & 55                         & 292     \\
      Yoruba                                          & Defoid                 & 318                        & 8,198   \\ 
      \bottomrule
    \end{tabular} 
  }
  \caption{Composition of the ``held-out'' language data (GCP = Greater Central Philippine).}
  \label{tab:data-held}
\end{table}

% \break

\section{Experimental setup} \label{sec:setup}

  We demonstrate multilabel morphosyntactic probing with multilingual BERT \cite[henceforth, \bert;][]{Devlin2018}, using morphologically annotated corpora from Universal Dependencies \cite[UD;][]{Nivre2016,Nivre2020}.\footnote{\url{https://universaldependencies.org/introduction.html}}

  \subsection{Data} \label{sec:data}

    Our target vectors draw on UD part of speech and morphological feature annotations. In a set of monolingual experiments, we trained separate probes to predict features from corpora for seven typologically-diverse languages: Afrikaans \cite[AfriBooms; cf.][]{Dirix2017}, Croatian \cite[SET; cf.][]{agic2015}, Finnish \cite[TDT; cf.][]{Haverinen2014,Pyysalo2015}, Hebrew \cite[HTB; cf.][]{Tsarfaty2013,McDonald2013,Sadde2018}, Korean \cite[PUD; cf.][]{Zeman2017}, Spanish \cite[AnCora; cf.][]{Alonso2016}, and Turkish \cite[IMST; cf.][]{Sulubacak2016,Tyers2017,Turk2019}.

    With the exception of the Korean data, all of the corpora came pre-split into training, validation, and test sets. We performed an 80-10-10 split on the 1,000-sentence Korean PUD corpus. Heeding conventional wisdom that says to throttle the training data for probes \cite{Zhang2018}, we reduced the other training sets to 800 sentences as well.

    Next, in a set of multilingual experiments, we trained probes on a shuffled combination of the training sentences from the monolingual probes. However, we excluded the Korean dataset from this analysis, due to the lack of documentation on its construction. The monolingual and multilingual datasets are summarized in Table \ref{tab:data-seen}.

    Finally, in a set of crosslingual transfer experiments, we evaluated the monolingual and multilingual probes on six held-out languages: Arabic \cite[PADT; cf.][]{Smrz2002,Smrz2008,Hajic2009}, Chinese \cite[PUD; cf.][]{Zeman2017}, Marathi \cite[UFAL; cf.][]{Ravishankar2017}, Slovenian \cite[SST; cf.][]{Dobrovoljc2016}, Tagalog (TRG), and Yoruba \cite[YTB; cf.][]{Ishola2020}. This data is summarized in Table \ref{tab:data-held}. It is important to note that \bert\ \emph{was} pre-trained on these languages; we only consider them ``held-out'' in that we never train probes to extract linguistic properties from these corpora (i.e., the experiments are zero-shot).

    All of the probes were trained to extract multiple features, such as part of speech, number, gender, case, and tense, as well as language-specific features, such as Finnish infinitive forms. (It is due to the inclusion of parts of speech that we refer to the proposed task as ``morpho\emph{-syntactic} tagging''.) Since the languages, and thus their corpora, often exhibit different morphosyntactic features, we use different feature label sets for each language and a semi-aggregated set for the multilingual probes. These feature label sets are listed in Appendix \ref{app:features}.

    Note that UD corpora assume an open vocabulary; many of the word types in the validation and test sets do not appear during training. This allows us to evaluate the effectiveness of the probes on out-of-vocabulary words. If the probes are truly extracting features versus memorizing the task, we would expect them to perform similarly on in-vocabulary and out-of-vocabulary words. (We touch on this in \S\ref{sec:hints}.) It is also worth noting that the UD corpora include decompositions of multiword tokens and separate annotations for their respective components. To keep the input to the probe faithful to naturalistic text (and to what \bert\ was presumably trained on), we embed the multiword tokens themselves, but aggregate the feature labels from their components (e.g., the Hebrew multiword token \<hspr> \emph{hasefer} `the book' is marked as both a determiner and noun).

  \subsection{Models, training, and implementation} \label{sec:models}

    For our experiments, we instantiated a ``BERT-Base, Multilingual Cased'' model using HuggingFace's \emph{Transformers} library \cite{Wolf2019b}. This BERT variant contains 110M parameters  across 12 Transformer layers, each with 12 attention heads and a hidden size of 768. The model was pre-trained on Wikipedia dumps from 104 languages. The authors oversampled the smaller Wikipedia corpora to create a more cross-linguistic vocabulary, consisting of 100K wordpieces.

    We froze \bert\ and trained linear classifiers on top of embeddings produced by \bert's initial embedding layer and each of its even-numbered Transformer layers \cite[cf.][]{Liu2019a,Hupkes2018}. Preliminary experiments showed that the even-numbered layers faithfully capture the layer-by-layer trends across \bert; we thus opted to cut down on compute by skipping the odd-numbered layers in our experiments. (We henceforth refer to the even-numbered layers as \bert-0, \bert-2, \bert-4, etc.)

    The classifiers used sigmoid activation and were trained with mean binary cross-entropy loss to perform the multilabel tagging task. We trained each classifier for 50 epochs, selecting the model from the epoch that achieved the best validation loss. Courtesy of PyTorch \cite{Paszke2019}, the classifiers were optimized using Adam \cite[learning rate = 0.001, $\beta_1$=0.9, $\beta_2$=0.999, $\epsilon$=1e-08;][]{Kingma2015}. No dropout was used.

    To conserve resources, we cached the sentence representations from \bert\ prior to training. These stored embeddings then served as inputs to the probing classifiers, in lieu of of passing a batch of input sentences through \bert\ during each training step. Since the probes themselves are simple linear layers and therefore non-contextual, this allowed us to batch the embeddings at the token level: We dispensed with the sequence length dimension and forewent padding. In all of the experiments, we opted for a batch size of 512 tokens (i.e., the batches had a dimensionality of 512$\times$768). This ``cache and batch'' approach allowed each monolingual probe to train in $\sim$1 minute and each multilingual probe in $\sim$4 minutes on a single Tesla K80 GPU. % (In fact, it took longer to embed the initial inputs and to obtain test predictions afterward than it did to train the probes.)

    Finally, despite BERT's word-piece vocabulary, we performed word-level predictions of morphosyntactic properties by aggregating the cached subword representations for each word, prior to training the probes. In exploratory experiments, we found that summing the subword representations per word performed best; we use this aggregation strategy throughout our experiments.\footnote{Summing the subword representations achieved comparable \fone\ scores but higher selectivity than taking their average.}

    \begin{figure*}[!ht] % mono
      \centering
      \includegraphics[width=\textwidth]{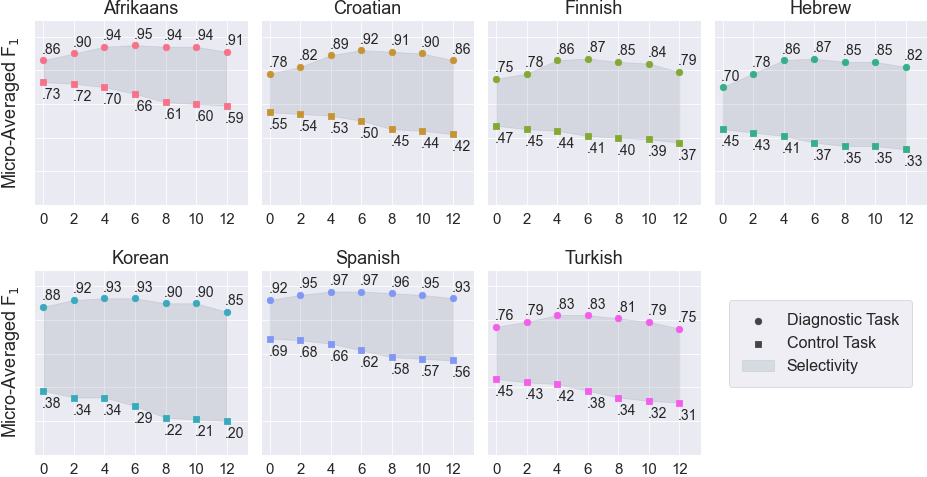}
      \caption{Micro-averaged \fone\ results from the monolingual probes on the diagnostic and control tasks. The $x$-axes indicate the \bert\ layer.}
      \label{fig:mono}
    \end{figure*}

  \subsection{Taking control} \label{sec:control}

    Following \citet{Hewitt2019b}, we constructed a control task to complement the multilabel tagging task, whereby each word type in the task vocabulary is assigned a multi-hot encoded vector that is randomly generated according to the true distribution of the feature labels in the training data. Deviating from Hewitt and Liang's notation, we generate a control output vector $\mathbf{c_i}$ for each word type $v_i \in V$,\footnote{$V$ is based on the word types across the training, dev, and test sets, since UD corpora use an open vocabulary.} such that $\mathbf{c}^i = c^i_1\ldots c^i_{|F|}$, where $c^i_j$ is sampled from the true distribution of feature $f_j$ in the training data. For instance, if $f_j$ is a feature of 4 out of 100 tokens in the training set, then $c^i_j$ has a 0.04 probability of being 1 for any word type $v_i$ (or, conversely, a 0.96 probability of being 0). To help ensure the presence of controlled counterparts for low-frequency features, we set a minimum probability threshold of 0.001. This generation procedure aims to have the likelihood of each feature in the gold data preserved in the control classification, allowing for a fairer comparison between the diagnostic and control tasks. In the control tasks for each language, the probes for each \bert\ layer were trained to predict the same control vectors.

\section{Monolingual experiments} \label{sec:mono}

  In a set of monolingual experiments, we trained and evaluated individual diagnostic probes on Afrikaans, Croatian, Finnish, Hebrew, Korean, Spanish, and Turkish, given representations from the even-numbered \bert\ layers. Their micro-averaged \fone\ scores are conveyed in Figure \ref{fig:mono}, along with their results on the analogous control tasks.

  \subsection{Monolingual performance at a glance}

    The micro-averaged \fone\ scores confirm that \bert\ renders many morphosyntactic properties easily extractable, with the best-performing probes for each language achieving scores between 0.83 and 0.97. We find that \bert-6 scored the highest  across the languages. This is consistent with prior work that has shown English BERT's middle layers to perform best on similar linguistic tasks \cite[][]{Liu2019a,Tenney2019}. Once \bert\ has encoded  morphologically relevant information, it seems that performance steadily declines in the topmost layers, as the representations gear up for cloze predictions.
    % CR! \cite{PLACEHOLDER}.

    Notably, the Afrikaans and Spanish probes performed the best and the Turkish probes the worst. It is tempting to conclude that `\bert\ knows Afrikaans and Spanish better than Turkish'. However, we should refrain from comparing probe performance across languages at the level of global scores, as each language differed in the sets of features that were extracted. Furthermore, although all of the probes were trained on 800 sentences, they were ultimately trained on varying numbers of tokens. It may not be a coincidence that the Afrikaans and Spanish probes performed the best and had the largest training sets, while Turkish had the smallest training set and the lowest \fone\ scores. 

  \subsection{Monolingual selectivity}

    While the diagnostic probes drastically outperformed their controlled counterparts,  we do see a trend of selectivity improving with the number of layers. This reinforces the findings of \citet{Hewitt2019b}, who posit that classifiers trained on top of lower layers are better equipped to memorize input-output mappings, due to their proximity to the initial vocabulary representations of the embedding layer. Nevertheless, the high selectivity scores across the probes show that a multilabel probing classifier offers a promising diagnostic of morphosyntactic representations.

    From a cross-linguistic standpoint, it is interesting that Afrikaans---the one morphologically \textit{impoverished} language in the bunch---exhibited the worst selectivity. This suggests that, perhaps, it is easier for probes to memorize mappings for analytic languages (i.e., languages that lack rich inflectional systems). However, we again mention that Afrikaans had the second largest training set, which could have given these probes relatively more opportunity to memorize the control task. Spanish, the language with the largest training set, displayed the second best performance on the control task.

    With the exception of Korean, we see that selectivity generally decreases as the amount of training data increases (cf. Table \ref{tab:size-sel}). Although this effect is somewhat surprising, given that each probe was trained on only 800 sentences, it is consistent with the idea that more training data will lead to more memorization \cite{Zhang2018} and, thus, higher \fone\ on the diagnostic and control tasks and lower selectivity \cite[cf.][]{Hewitt2019b}. These findings suggest that token-level probes are highly sensitive to their number of training tokens, and that the traditional method of controlling for dataset size by limiting the number of \emph{sentences} may be insufficient for fair comparisons between probes. Future work should explore controlling for the number of \emph{tokens} that appear during training and its effects on selectivity. This is made straightforward by caching the embeddings and de-sequencing the inputs, as described in \S\ref{sec:models}.

  \subsection{Case study: Hebrew covert determiners} \label{sec:heb-art}

    The micro-averaged scores in Figure \ref{fig:mono} show that \bert\ has indeed learned \emph{some} linguistic system or portion thereof. However, these scores do not give much insight into which aspects of morphosyntax \bert\ has come to represent, the interplay between these properties, nor how much \bert\ varies in capturing each feature label. Crucially, a key strength of multilabel probing is that it makes it easy to mine fine-grained morphosyntactic observations that implicate multiple features. In this section, we present such an analysis with Hebrew determiners, inspired by \citet{Klein2020}. We focus on the predictions from \bert-6, since it displayed the highest F$_1$ and selectivity scores out of the Hebrew probes.

    \begin{table}[t] % training size v. selectivity
      \centering
      \begin{tabular}{lrrl}
        \toprule
          Language    & Train Size    & Sel.   & \bert- \\
        \midrule
          Spanish     & 24,345    & 0.38   & 8, 10    \\
          Afrikaans   & 21,160    & 0.34   & 10       \\
          Croatian    & 17,811    & 0.46   & 8, 10    \\
          Hebrew      & 16,061    & 0.50   & 6, 8, 10 \\
          Korean      & 13,177    & 0.69   & 10       \\
          Finnish     & 10,786    & 0.46   & 6        \\
          Turkish     & 9,768     & 0.47   & 8, 10    \\
        \bottomrule
      \end{tabular}
      \caption{Training set size (in tokens) and the highest monolingual selectivity (Sel.) achieved per language.}
      \label{tab:size-sel}
    \end{table}

    \begin{table}[t] % Hebrew DET
      \centering
      \begin{tabular}{lrrr}
        \toprule
          \texttt{PronType=Art}  & P    & R    & \fone \\
        \midrule
          Overt determiner       & 0.93 & 0.56 & 0.70 \\
          Covert determiner      & 0.69 & 0.40 & 0.50 \\
        \bottomrule
      \end{tabular}
      \caption{Recognition of the feature \texttt{PronType=Art} in ADP-DET-NOUN multiword tokens, given the Hebrew \bert-6 probe.}
      \label{tab:heb-art}
    \end{table}

    \begin{figure*}[!h] % multi
      \centering
      \includegraphics[width=\textwidth]{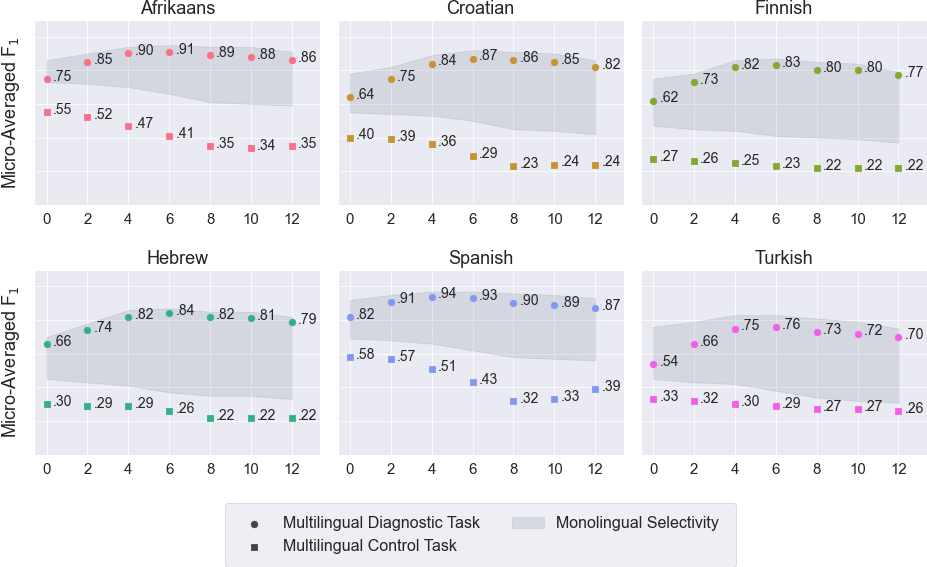}
      \caption{Micro-averaged \fone\ results from the multilingual probes on the diagnostic and control tasks for each language. The $x$-axes indicate the \bert\ layer. The depicted monolingual results (for comparison) assume the same feature label subsets as the multilingual models; incidentally, the monolingual diagnostic task scores are equivalent to the scores reported in Figure \ref{fig:mono}, while the control task scores differ by $\pm 2$ points.} 
      \label{fig:multi}
    \end{figure*}

    Ambiguous orthographies as well as multiword tokens (MWTs) are ubiquitous in Hebrew. As stated previously, we represented MWTs by flattening their structure and labeling each MWT with the feature labels of its components. A common structure of MWTs in Hebrew is ADP-(DET)-NOUN, where the determiner is the definite article -\<h> \emph{ha} `the'. Depending on the preposition, the definite article is represented orthographically (e.g., -\<mh> \emph{miha} `from the') or as a vowel change on the preposition that is not represented orthographically (e.g., -\<l> can be either \emph{le} `to a' or \emph{la} `to the'). When the article is absent from the orthography, we refer to it as being \emph{covert}.\footnote{Note that this is a slightly different usage of \emph{covert} than in generative syntactic theory.}

    The definite article is a subset of the determiners in the HTB corpus, but is uniquely identified by the label \texttt{PronType=Art}. We thus extracted all of the ADP-(DET)-NOUN cases from the Hebrew test set (234 in total), and examined how well \bert-6 captured this property. Overall, we found that it was less able to recognize \texttt{PronType=Art} when the article was not overt (Table \ref{tab:heb-art}).

    Yet, we also found that agreement patterns facilitated recognition of the covert definite article. In particular, Hebrew adjectival modifiers agree with the nouns they modify in gender, number, and definiteness (e.g., in the noun phrase \<\underline{h}byt \underline{h}q.tN> \emph{\underline{ha}bayit \underline{ha}katan} `the small house', \<byt> \emph{bayit} is `house.sing.masc' and \<q.tN> \emph{katan} is `small.sing.masc'). Based on UD's \texttt{amod} annotations, MWTs that appeared in these constructions constituted 44.3\% of TPs, 19.4\% of FPs, and 26.2\% of FNs when identifying the covert definite article. Moreover, the majority of the FNs involved additional erroneous predictions, where either \texttt{PronType=Art} was not captured on the modifier, the parts of speech were misidentified, or the modifier and the noun were mis-predicted to disagree along an additional feature (i.e., gender or number). These concomitant errors were largely missing from the TPs. 

    It seems that \bert-6 has learned that Hebrew nouns and their modifiers agree along multiple features, and that it is able to use the presence of an overt definite article on a modifier to help infer the presence of a covert article in a MWT. When not all of the grammatical features that participate in agreement are captured, this can attenuate recognition of the covert article (and vice versa).

\section{Multilingual experiments} \label{sec:multi}

  So far, we have used monolingual probes to assess the linguistic representations from multilingual BERT on a language-by-language basis. One question pertaining to efficiency remains: Can we replace the individual monolingual probes with a single multilingual probe and derive comparable insights? To address this question, we additionally trained multilingual probes on a shuffled combination of the training sets for Afrikaans, Croatian, Finnish, Hebrew, Spanish, and Turkish. The multilingual probes extracted an aggregated subset of the features captured by the monolingual probes. We then assessed the multilingual probes' performance on each language independently. Overall, the multilingual probes exhibited slight dips in performance, but better selectivity, compared to their monolingual counterparts (Figure \ref{fig:multi}). These trends occurred despite all of the multilingual models converging before they reached epoch 50 during training.

  \subsection{Multilingual task complexity} \label{sec:multi-task}

    Even though the multilingual experiments merely combine the monolingual training data, the multilingual task is inherently more complex than the monolingual task. In particular, the probes must balance the needs of multiple languages and extract features from a broader diversity of data.

    Consider nominative case. When focusing on predictions from \bert-6, we see that the \texttt{Case=Nom} scores for each language dipped with the multilingual probe (Table \ref{tab:case-nom}). Importantly, the distribution of nominative case morphology differs across languages; for instance, nominative case only appears on pronouns in Spanish, whereas it can appear on nouns, verbs, and adjectives in Turkish (according to the AnCora and IMST corpora). It is possible that such variation might result in ``conflicting'' training signals to the probe, causing the dips in the multilingual probe's performance. Furthermore, it suggests that, although \bert\ renders nominative case easily extractable for each language independently, \bert\ has not recognized this morphology to correspond to the same nominative notion. We return to this point in \S\ref{sec:transfer}.

    \begin{table}[t]  % nominative case
      \centering
      \begin{tabular}{lrrrrr}
        \toprule
          Probe         & Af    & Hr    & Fi    & Es    & Tr \\
        \midrule
          \emph{Mono.}  & 0.89  & 0.88  & 0.89  & 0.96  & 0.79 \\
          \emph{Multi.} & 0.71  & 0.76  & 0.80  & 0.14  & 0.65 \\
        \bottomrule
      \end{tabular}
      \caption{\fone\ results for nominative case (\texttt{Case=Nom}) in Afrikaans (Af), Croatian (Hr), Finnish (Fi), Spanish (Es), and Turkish (Tr), given the monolingual and multilingual \bert-6 probes.}
      \label{tab:case-nom}
    \end{table}

    The challenge posed by multilingual multilabel tagging signals the possibility that a linear probe is not complex enough to accommodate the task. In a small set of exploratory experiments that replaced the linear classifier with a multilayer perceptron, we found that this boosted the multilingual probes' performance, but at the expense of selectivity. As the number of parameters increased, the micro-averaged \fone\ performance would approach that of the monolingual probes, but with comparable or worse selectivity. (See Appendix \ref{app:mlp} for details of this analysis.) This suggests that the improvements observed by the more complex probes result from memorization \cite[cf.][]{Hewitt2019b}.

  \subsection{Hints of memorization} \label{sec:hints}

    Indeed, another potential explanation for the contrast in monolingual and multilingual performance is that the simpler task affords the monolingual probes more opportunity to memorize the feature labels. This explanation is supported by how the multilingual probes generally exhibit greater selectivity and accounts for why their performance deficit is, for the most part, spread evenly across the feature labels (see Appendix \ref{app:perf}).

    If the monolingual probes do rely more on memorization, this would predict that the multilingual probes are better able to generalize to new data, by depending more on the input representations. This prediction is largely borne out with out-of-vocabulary (OOV) tokens: We micro-averaged separate \fone\ scores for the words that were seen during training and those that weren't, the intuition being that a probe that generalizes better is more extractive and will exhibit smaller gaps in performance between OOV and in-vocabulary (IV) words. For Croatian, Finnish, Hebrew, and Turkish, we observed that the gaps between IV and OOV performance tended to be smaller for the multilingual probes than the monolingual models, especially in later layers.\footnote{In contrast, for Spanish, the multilingual probes generally exhibited greater IV-OOV gaps than the monolingual models, though this trend diminished with the number of layers. Likewise, for Afrikaans, the IV-OOV gaps were very similar between the monolingual and multilingual probes. Crucially, relative to the other languages, the IV-OOV gaps were greatest for Spanish and Afrikaans (where IV performance was better) in both the monolingual and multilingual settings. This reversal of trends is likely due to their substantially larger trainings sets: The increased number of training tokens (and also training steps) lured the multilingual probes to memorize the word-to-label mappings for these languages.} Impressively, some of the Hebrew and Turkish probes performed \emph{better} amongst OOV tokens than with IV tokens, as previewed in Figure \ref{fig:iv-oov-small}. Fully visualized in Appendix \ref{app:oov}, these trends further suggest that the monolingual probes are more inclined towards memorization than the multilingual probes.

    \begin{figure}[t] % iv - oov (small)
      \centering
      \includegraphics[width=20em]{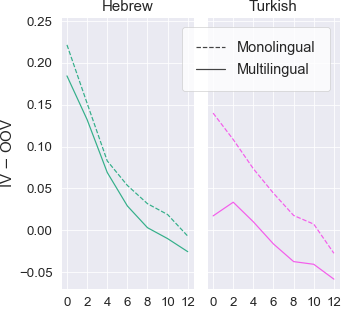}
      \caption{Generalizability of the Hebrew and Turkish monolingual and multilingual probes; calculated by subtracting the micro-averaged OOV \fone\ scores from the micro-averaged IV \fone\ scores. The $x$-axes indicate the \bert\ layer.}
      \label{fig:iv-oov-small}
    \end{figure}

    \begin{figure*}[t] % transfer
      \centering
      \includegraphics[width=\textwidth]{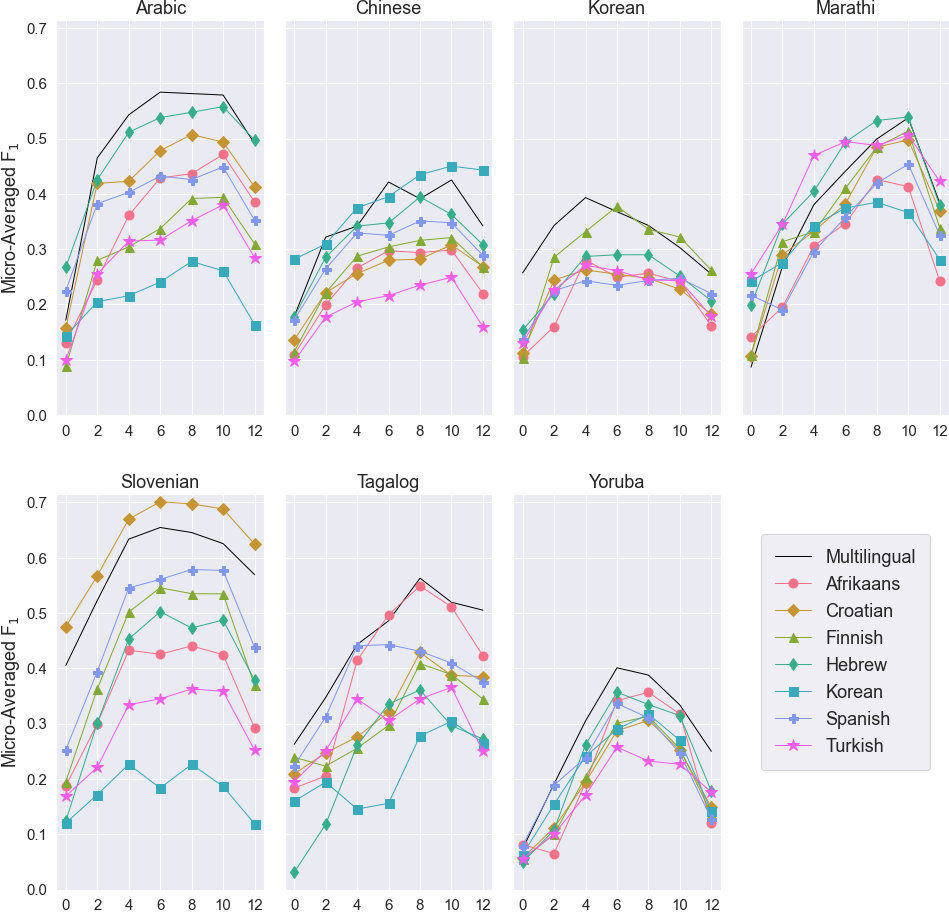}
      \caption{Micro-averaged \fone\ results from evaluating the monolingual and multilingual probes on the ``held-out'' languages (plus Korean). The $x$-axes indicate the \bert\ layer.}
      \label{fig:transfer}
    \end{figure*}

    Another piece of evidence comes from language-specific features. For the multilingual experiments, we included two sets of language-specific features: Finnish infinitive forms and Hebrew verb classes (a.k.a. \emph{binyanim}). While the monolingual probes generally outperformed their multilingual counterparts at the feature level, the opposite tended to be true for language-specific features (see Appendices \ref{app:perf-fi} and \ref{app:perf-he}). If the multilingual probes are more extractive, especially with cross-linguistic features, this might leave the probe with more ``room'' to capture language-specific features, either through memorization or extraction. % Base forms are an exception.

\begin{figure*}[t] % transfer-ex
  \centering
  \includegraphics[width=\textwidth]{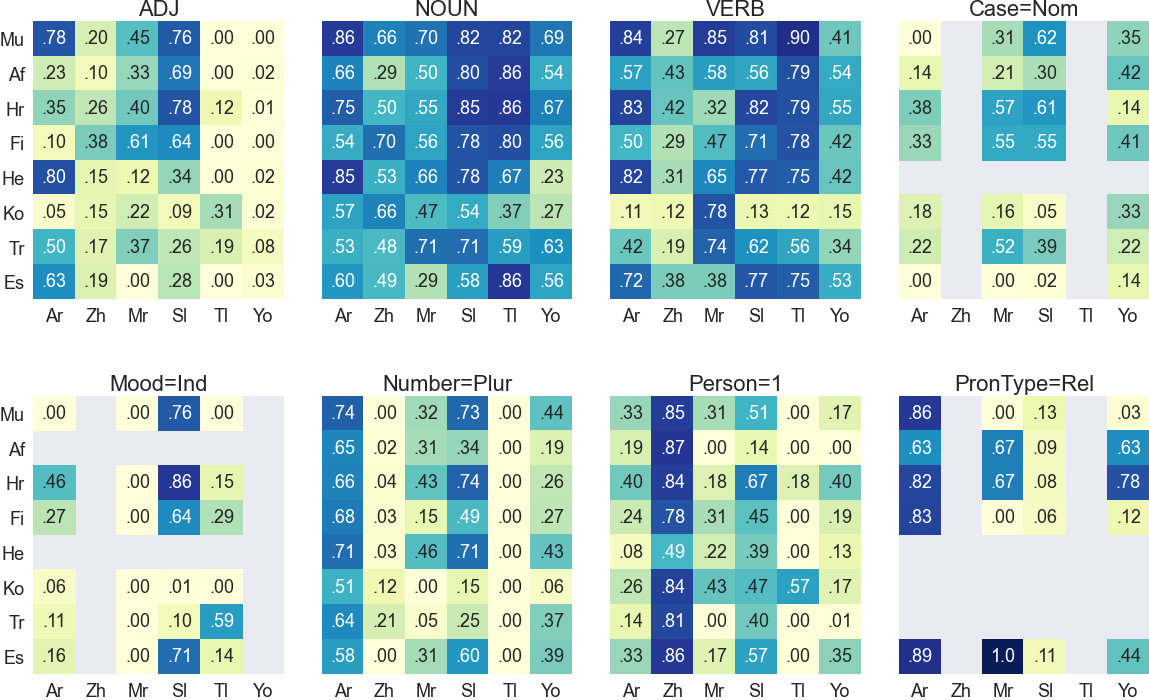}
  \caption{A handful of feature-level \fone\ results from evaluating the monolingual and multilingual \bert-6 probes on ``held-out'' languages. The $x$-axes indicate the held-out language (Ar=Arabic, Zh=Chinese,  Mr=Marathi, Sl=Slovenian, Tl=Tagalog, and Yo=Yoruba), while the $y$-axes indicate the probe (Mu=Multilingual, Af=Afrikaans, Hr=Croatian, Fi=Finnish, He=Hebrew, Ko=Korean, Es=Spanish, and Tr=Turkish). Grayed-out regions indicate where the feature is not applicable to the language or annotated in the language's corpus. In a fun example, the Marathi test set contained exactly one relative pronoun (\texttt{PronType=Rel}), which was perfectly identified by the Spanish probe (\fone\ $= 1.0$).}
  \label{fig:transfer-ex}
\end{figure*}

\section{Crosslingual experiments} \label{sec:transfer}

  We have shown that multilabel probing offers a robust method for analyzing which morphosyntactic features are easily extractable from language model embeddings. However, these probes can also be used to study which morphosyntactic features are encoded similarly cross-linguistically. In particular, when a monolingual probe successfully extracts a feature label from a held-out language, this suggests that the \nlm\ has come to recognize this property as being shared by the two languages. 

  In this section, we evaluate the monolingual and multilingual probes on the UD test sets for Arabic, Chinese, Marathi, Slovenian, Tagalog, and Yoruba. These experiments are akin to prior work on zero-shot crosslingual transfer \cite{Pires2019,Wu2020,Conneau2020b,K2020}, though we differ in that we never fine-tune \bert\ on our ``source'' languages. Figure \ref{fig:transfer} shows the monolingual and multilingual micro-averaged \fone\ performance across the held-out languages. Focusing once more on \bert-6, we also examine a small subset of labels, presented in Figure \ref{fig:transfer-ex}.

  \subsection{Towards cross-linguistic categories}

    Overall, we find that the probes performed relatively well on extracting nouns and verbs (to a slightly lesser extent) across the held-out languages. This suggests that \bert\ encodes \emph{noun}-hood and \emph{verb}-hood in a cross-linguistic fashion---that it has some conception of nouns and verbs that transcends individual languages.

    \emph{Adjective}-hood, in contrast, seems to be represented less cohesively. All of the probes struggled to identify adjectives in Chinese, and even more so in Tagalog and Yoruba. This is not to say that \bert\ does not capture the distribution of adjectives in these languages, but, rather, that it has not connected them to their counterparts in other languages. This may be especially true of low-resource languages like Tagalog and Yoruba.\footnote{Recall that \bert\ was trained the languages with the top 100 largest Wikipedias. Based on Wikimedia's \href{https://meta.wikimedia.org/wiki/List_of_Wikipedias}{\emph{List of Wikipedias}}, it seems that the Wikipedia dumps for Tagalog and Yoruba were among the smallest corpora that \bert\ was trained on, ranking 92 and 106 at present, respectively. Note also that, in the ``language resource race'', \citet{Joshi2020} give Tagalog and Yoruba scores of 3/5 and 2/5.} Even though \bert's training involved over-sampling smaller corpora, it might be the case that a multilingual \nlm\ requires more exposure to adjectives in a wider array of contexts from a given language in order to relate them to their counterparts cross-linguistically (see \citealt{Conneau2020a} for an interesting relevant discussion).

    Cross-linguistic variation in a feature's distribution in natural languages might also lead a \nlm\ not to recognize when a property is shared by multiple languages. In \S\ref{sec:multi-task}, we cited such variation as the reason the multilingual probes struggled with nominative case, positing that \bert\ had not recognized nominative case morphology in different languages to correspond to the same nominative notion. We see this suspicion borne out in Figure \ref{fig:transfer-ex}, where predictions of \texttt{Case=Nom} in the held-out languages ranged from 0 to 0.62 \fone. As evidenced by the lack of nominative transfer, it may be that cross-linguistic variation in the distribution of nominative morphology led to a decentralized encoding of nominative case in \bert; as a result, this made it more challenging for the multilingual probes to capture nominative case. In other words, the highly contrasted representations of nominative case from the different languages acted as ``conflicting'' training signals to the multilingual probes.

    However, there are a handful of cases where the multilingual probes performed \emph{best}. Most strikingly, the \bert-6 multilingual probe obtained 0.90 \fone\ on Tagalog verbs, whereas none of the monolingual probes got over 0.79 \fone. This suggests that, in the absence of conflicting representations and with linguistic properties that are encoded more cohesively, such as \emph{verb}-hood, exposure to multiple languages can lead a probe to forge more replete connections with \bert's representational space when extracting those properties.

  \subsection{Family ties}

    In the absence of truly cross-linguistic representations, we generally find that a  monolingual probe extends equivalently or better to a held-out language than the multilingual model. In particular, the monolingual probes often faired well with \emph{related} languages \cite[cf.][]{Pires2019,Wu2020,Conneau2020b}. For instance, compared to the other monolingual probes, the Hebrew probes faired best with Arabic, another Semitic language, topping out at a micro-averaged \fone\ score of 0.56 (\bert-10). This was also the case at the feature level with nouns, verbs, adjectives, and plurals, as shown in Figure \ref{fig:transfer-ex}. Notably, Hebrew and Arabic use different scripts; therefore, if \bert\ has come to represent them similarly, this likely falls out of structural similarities between the two languages.

    Likewise, the Croatian \bert-6 probe got a micro-averaged \fone\ score of 0.70 on Slovenian. (For comparison, the Turkish \bert-6 probe scored 0.76 \fone\ \emph{on Turkish}.) In Figure \ref{fig:transfer-ex}, the Croatian probe also performed the best on Slovenian nouns, verbs, adjectives, and words inflected for first person, plurality, or indicative mood. This success seems due to both structural and surface similarities (e.g., cognates) between Croatian and Slovenian. For instance, Croatian achieves 0.95 \fone\ on conditional mood (\texttt{Mood=Cnd}; see Appendix \ref{app:cross}) and 0.86 \fone\ on indicative mood in Slovenian because the two languages share several auxiliaries that mark mood (e.g., \emph{bi} for conditional, \emph{je} for indicative).

  \subsection{Revisiting memorization}

    Note that, with the exception of shared morphemes, the successful instances of crosslingual transfer cannot be reduced to memorization. If the probes merely memorized their monolingual training data, one would expect chance performance and less variability when evaluating them on the held-out languages. These evaluations  further verify that our multilabel probes extract meaningful representations from \bert. In addition, when applied to held-out languages, they provide a supplementary method for gauging the complexity of a probe and its ability to memorize a linguistic task.
 
\section{Discussion} \label{sec:discussion}

  We have developed a multilabel probing task to assess the morphosyntactic representations of contextualized word embeddings from multilingual language models. We demonstrated this task with multilingual BERT \cite[fondly known as \bert;][]{Devlin2018}, devoting special attention to 13 typologically diverse languages of varying morphological complexity: Afrikaans, Arabic, Chinese, Croatian, Finnish, Hebrew, Korean, Marathi, Slovenian, Spanish, Tagalog, Turkish, and Yoruba. Through this minimally complex but robust paradigm, we have shown that \bert\ renders many morphosyntactic features easily and simultaneously extractable. 

  % monolingual
  In a set of monolingual experiments (\S\ref{sec:mono}), we found that \bert-6 contained the most morphosyntactic information, with the probes obtaining micro-averaged \fone\ scores between 0.83 and 0.97 and selectivity scores between 0.29 and 0.64. In a small case study of Hebrew determiners (\S\ref{sec:heb-art}), we illustrated an analysis that implicates \emph{multiple} features (i.e., lexical category, pronominal type, number, and gender). Crucially, traditional probing methods that treat recognition of a feature as its own task would requiring training multiple models to arrive at the same analysis. In this way, the multilabel training objective is both more efficient and more informative.  

  % multilingual (+ memorization)
  Next, in a set of multilingual experiments (\S\ref{sec:multi}), we saw that the multilingual probes marginally underperformed their monolingual counterparts, while largely upholding the same trends and exhibiting better selectivity. We attributed their contrast in performance to the monolingual probes relying more on memorization, given a simpler task (\S\ref{sec:hints}). In sum, these findings indicate that the multilingual probes may be more trustworthy or ``expressive'' diagnostics of linguistic representations \cite[cf.][]{Hewitt2019b}. However, since our goal is to probe embeddings rather than to perform state-of-the-art morphosyntactic tagging, the monolingual and multilingual probes offer the same \emph{insights} to the extent that they exhibit comparable trends and lend themselves to the same generalizations.

  % crosslingual
  Furthermore, while the multilingual probes are more efficient and less inclined towards memorization, they are also more susceptible to crosslingual interference when the \nlm\ has encoded the same property differently across languages (such as \bert\ with nominative case). In a set of crosslingual transfer experiments, we further evaluated the monolingual and multilingual probes on data from ``held-out'' languages (\S\ref{sec:transfer}). We showed that applying this probing paradigm accordingly can illuminate which linguistic properties a \nlm\ has come to recognize as being shared across languages. In particular, we found that \bert\ does not always hold cohesive representations of a particular feature cross-linguistically. We conjectured that multilingual probes more reliably extract linguistic properties that are encoded similarly cross-linguistically, or that are alternatively language-specific.

\section{Conclusion} \label{sec:conclusion}

  Emerging studies on interpretability have highlighted a wealth of linguistic information that can be extracted from neural language model embeddings  \cite[e.g.,][]{Conneau2018a,Gulordava2018,Hupkes2018,Marvin2018,Zhang2018,Bacon2019,Futrell2019,Hewitt2019a,Jawahar2019,Liu2019a,Sahin2020,Tenney2019,Chi2020,Edmiston2020}. Our paper contributes to this effort by providing an efficient way to probe dozens of morphosyntactic properties simultaneously; each of our probes extracted 35-89 feature labels and trained in approximately 1-4 minutes on a single K80 GPU. Predictions from these probes can be used to perform holistic analyses of a multilingual model's ability to encode systems of morphology, as well as more fine-grained analyses of individual features, agreement phenomena, and and how shared properties are represented cross-linguistically. 

  % future work
  % \break
  We encourage future probing efforts to take advantage of the multilabel probing paradigm, which yields rich insights at low costs. We plan to release our ``predictions datasets'' to support more detailed analyses of \bert's facility for morphosyntax; our work can also be used to focus future contributions by identifying which \bert\ layers to target for more in-depth probing of a particular feature. Finally, future research should explore how global- and, specifically, feature-level performance corresponds to the performance of downstream multilingual systems, especially amongst morphologically rich languages.

\section*{Acknowledgments}
  
  We thank the UW Research Computing Club for supporting our research through their Cloud Credit Program and Fei Xia for her helpful feedback and discussions.

\bibliography{bibliography}
\bibliographystyle{acl_natbib}

\newpage
\onecolumn
\begin{appendices}

  \section{Morphosyntactic feature labels} \label{app:features}  % A

    We considered 166 morphosyntactic feature labels in total across our experiments. The monolingual probes were trained to extract every morpho-syntactically relevant feature that was available for the given language in its UD corpus. \\

    \rowcolors{1}{white}{whitesmoke}
    \begin{longtable}{lC{1.25cm}C{1.25cm}C{1.25cm}C{1.25cm}C{1.25cm}C{1.25cm}C{1.25cm}}
      \caption*{Table A1: The monolingual probes extracted different sets of features, while the multilingual probes extracted a semi-aggregated subset of these features (in bold under ``Feature Labels'').} \\
      \toprule\hiderowcolors
        Feature Labels               &   Afrikaans &    Croatian &     Finnish &      Hebrew &      Korean &     Spanish &     Turkish \\
      \midrule\showrowcolors
        \textbf{ADJ}                 &  \checkmark &  \checkmark &  \checkmark &  \checkmark &  \checkmark &  \checkmark &  \checkmark \\
        \textbf{ADP}                 &  \checkmark &  \checkmark &  \checkmark &  \checkmark &             &  \checkmark &  \checkmark \\
        \textbf{ADV}                 &  \checkmark &  \checkmark &  \checkmark &  \checkmark &  \checkmark &  \checkmark &  \checkmark \\
        \textbf{AUX}                 &  \checkmark &  \checkmark &  \checkmark &  \checkmark &  \checkmark &  \checkmark &  \checkmark \\
        \textbf{CCONJ}               &  \checkmark &  \checkmark &  \checkmark &  \checkmark &  \checkmark &  \checkmark &  \checkmark \\
        \textbf{DET}                 &  \checkmark &  \checkmark &             &  \checkmark &  \checkmark &  \checkmark &  \checkmark \\
        \textbf{NOUN}                &  \checkmark &  \checkmark &  \checkmark &  \checkmark &  \checkmark &  \checkmark &  \checkmark \\
        \textbf{NUM}                 &  \checkmark &  \checkmark &  \checkmark &  \checkmark &  \checkmark &  \checkmark &  \checkmark \\
        \textbf{PART}                &  \checkmark &  \checkmark &             &             &  \checkmark &  \checkmark &             \\
        \textbf{PRON}                &  \checkmark &  \checkmark &  \checkmark &  \checkmark &  \checkmark &  \checkmark &  \checkmark \\
        \textbf{PROPN}               &  \checkmark &  \checkmark &  \checkmark &  \checkmark &  \checkmark &  \checkmark &  \checkmark \\
        \textbf{SCONJ}               &  \checkmark &  \checkmark &  \checkmark &  \checkmark &             &  \checkmark &             \\
        \textbf{VERB}                &  \checkmark &  \checkmark &  \checkmark &  \checkmark &  \checkmark &  \checkmark &  \checkmark \\
        AdjType=Attr                 &  \checkmark &             &             &             &             &             &             \\
        AdjType=Pred                 &  \checkmark &             &             &             &             &             &             \\
        AdpType=Post                 &             &             &  \checkmark &             &             &             &             \\
        AdpType=Prep                 &  \checkmark &             &  \checkmark &             &             &  \checkmark &             \\
        AdpType=Preppron             &             &             &             &             &             &  \checkmark &             \\
        AdvType=Tim                  &             &             &             &             &             &  \checkmark &             \\
        Animacy=Anim                 &             &  \checkmark &             &             &             &             &             \\
        Animacy=Inan                 &             &  \checkmark &             &             &             &             &             \\
        Aspect=Hab                   &             &             &             &             &             &             &  \checkmark \\
        Aspect=Perf                  &             &             &             &             &             &             &  \checkmark \\
        Aspect=Prog                  &             &             &             &             &             &             &  \checkmark \\
        Aspect=Prosp                 &             &             &             &             &             &             &  \checkmark \\
        Aspect=Rapid                 &             &             &             &             &             &             &  \checkmark \\
        \textbf{Case=Abe}            &             &             &  \checkmark &             &             &             &             \\
        \textbf{Case=Abl}            &             &             &  \checkmark &             &             &             &  \checkmark \\
        \textbf{Case=Acc}            &  \checkmark &  \checkmark &  \checkmark &  \checkmark &  \checkmark &  \checkmark &  \checkmark \\
        \textbf{Case=Ade}            &             &             &  \checkmark &             &             &             &             \\
        Case=Advb                    &             &             &             &             &  \checkmark &             &             \\
        \textbf{Case=All}            &             &             &  \checkmark &             &             &             &             \\
        \textbf{Case=Com}            &             &             &  \checkmark &             &             &  \checkmark &             \\
        Case=Comp                    &             &             &             &             &  \checkmark &             &             \\
        \textbf{Case=Dat}            &             &  \checkmark &             &             &             &  \checkmark &  \checkmark \\
        \textbf{Case=Ela}            &             &             &  \checkmark &             &             &             &             \\
        \textbf{Case=Equ}            &             &             &             &             &             &             &  \checkmark \\
        \textbf{Case=Ess}            &             &             &  \checkmark &             &             &             &             \\
        \textbf{Case=Gen}            &             &  \checkmark &  \checkmark &  \checkmark &  \checkmark &             &  \checkmark \\
        \textbf{Case=Ill}            &             &             &  \checkmark &             &             &             &             \\
        \textbf{Case=Ine}            &             &             &  \checkmark &             &             &             &             \\
      \pagebreak\hiderowcolors
      \multicolumn{8}{l}{\emph{Continuation of Table A1:}} \\ \\
      \toprule
        Feature Labels               &   Afrikaans &    Croatian &     Finnish &      Hebrew &      Korean &     Spanish &     Turkish \\
      \midrule\showrowcolors
        \textbf{Case=Ins}            &             &  \checkmark &  \checkmark &             &             &             &  \checkmark \\
        \textbf{Case=Loc}            &             &  \checkmark &             &             &             &             &  \checkmark \\
        \textbf{Case=Nom}            &  \checkmark &  \checkmark &  \checkmark &             &  \checkmark &  \checkmark &  \checkmark \\
        \textbf{Case=Par}            &             &             &  \checkmark &             &             &             &             \\
        \textbf{Case=Tem}            &             &             &             &  \checkmark &             &             &             \\
        \textbf{Case=Tra}            &             &             &  \checkmark &             &             &             &             \\
        \textbf{Case=Voc}            &             &  \checkmark &             &             &             &             &             \\
        Clitic=Han                   &             &             &  \checkmark &             &             &             &             \\
        Clitic=Ka                    &             &             &  \checkmark &             &             &             &             \\
        Clitic=Kaan                  &             &             &  \checkmark &             &             &             &             \\
        Clitic=Kin                   &             &             &  \checkmark &             &             &             &             \\
        Clitic=Ko                    &             &             &  \checkmark &             &             &             &             \\
        Clitic=Pa                    &             &             &  \checkmark &             &             &             &             \\
        Clitic=S                     &             &             &  \checkmark &             &             &             &             \\
        Connegative=Yes              &             &             &  \checkmark &             &             &             &             \\
        Definite=Cons                &             &             &             &  \checkmark &             &             &             \\
        Definite=Def                 &  \checkmark &  \checkmark &             &  \checkmark &             &  \checkmark &             \\
        Definite=Ind                 &  \checkmark &  \checkmark &             &             &             &  \checkmark &             \\
        Degree=Abs                   &             &             &             &             &             &  \checkmark &             \\
        Degree=Cmp                   &  \checkmark &  \checkmark &  \checkmark &             &             &  \checkmark &             \\
        Degree=Dim                   &  \checkmark &             &             &             &             &             &             \\
        Degree=Pos                   &  \checkmark &  \checkmark &  \checkmark &             &             &             &             \\
        Degree=Sup                   &  \checkmark &  \checkmark &  \checkmark &             &             &  \checkmark &             \\
        Derivation=Inen              &             &             &  \checkmark &             &             &             &             \\
        Derivation=Ja                &             &             &  \checkmark &             &             &             &             \\
        Derivation=Lainen            &             &             &  \checkmark &             &             &             &             \\
        Derivation=Llinen            &             &             &  \checkmark &             &             &             &             \\
        Derivation=Minen             &             &             &  \checkmark &             &             &             &             \\
        Derivation=Sti               &             &             &  \checkmark &             &             &             &             \\
        Derivation=Tar               &             &             &  \checkmark &             &             &             &             \\
        Derivation=Ton               &             &             &  \checkmark &             &             &             &             \\
        Derivation=Ttain             &             &             &  \checkmark &             &             &             &             \\
        Derivation=U                 &             &             &  \checkmark &             &             &             &             \\
        Derivation=Vs                &             &             &  \checkmark &             &             &             &             \\
        Echo=Rdp                     &             &             &             &             &             &             &  \checkmark \\
        Evident=Nfh                  &             &             &             &             &             &             &  \checkmark \\
        Form=Adn                     &             &             &             &             &  \checkmark &             &             \\
        Form=Aux                     &             &             &             &             &  \checkmark &             &             \\
        Form=Compl                   &             &             &             &             &  \checkmark &             &             \\
        \textbf{Gender=Fem}          &             &  \checkmark &             &  \checkmark &             &  \checkmark &             \\
        \textbf{Gender=Masc}         &             &  \checkmark &             &  \checkmark &             &  \checkmark &             \\
        \textbf{Gender=Neut}         &             &  \checkmark &             &             &             &             &             \\
        Gender[psor]=Fem             &             &  \checkmark &             &             &             &             &             \\
        Gender[psor]=Masc            &             &  \checkmark &             &             &             &             &             \\
        Gender[psor]=Neut            &             &  \checkmark &             &             &             &             &             \\
        \textbf{HebBinyan=HIFIL}     &             &             &             &  \checkmark &             &             &             \\
        \textbf{HebBinyan=HITPAEL}   &             &             &             &  \checkmark &             &             &             \\
        \textbf{HebBinyan=HUFAL}     &             &             &             &  \checkmark &             &             &             \\
      \pagebreak\hiderowcolors
      \multicolumn{8}{l}{\emph{Continuation of Table A1:}} \\ \\
      \toprule
        Feature Labels               &   Afrikaans &    Croatian &     Finnish &      Hebrew &      Korean &     Spanish &     Turkish \\
      \midrule\showrowcolors
        \textbf{HebBinyan=NIFAL}     &             &             &             &  \checkmark &             &             &             \\
        \textbf{HebBinyan=PAAL}      &             &             &             &  \checkmark &             &             &             \\
        \textbf{HebBinyan=PIEL}      &             &             &             &  \checkmark &             &             &             \\
        \textbf{HebBinyan=PUAL}      &             &             &             &  \checkmark &             &             &             \\
        \textbf{HebExistential=True} &             &             &             &  \checkmark &             &             &             \\
        \textbf{InfForm=1}           &             &             &  \checkmark &             &             &             &             \\
        \textbf{InfForm=2}           &             &             &  \checkmark &             &             &             &             \\
        \textbf{InfForm=3}           &             &             &  \checkmark &             &             &             &             \\
        \textbf{Mood=Cnd}            &             &  \checkmark &  \checkmark &             &             &  \checkmark &  \checkmark \\
        Mood=Des                     &             &             &             &             &             &             &  \checkmark \\
        Mood=Gen                     &             &             &             &             &             &             &  \checkmark \\
        \textbf{Mood=Imp}            &             &  \checkmark &  \checkmark &  \checkmark &  \checkmark &  \checkmark &  \checkmark \\
        \textbf{Mood=Ind}            &             &  \checkmark &  \checkmark &             &  \checkmark &  \checkmark &  \checkmark \\
        Mood=Nec                     &             &             &             &             &             &             &  \checkmark \\
        Mood=Opt                     &             &             &             &             &             &             &  \checkmark \\
        Mood=Pot                     &             &             &  \checkmark &             &             &             &  \checkmark \\
        Mood=Sub                     &             &             &             &             &             &  \checkmark &             \\
        NumType=Card                 &             &  \checkmark &  \checkmark &             &  \checkmark &  \checkmark &  \checkmark \\
        NumType=Dist                 &             &             &             &             &             &             &  \checkmark \\
        NumType=Frac                 &             &             &             &             &             &  \checkmark &             \\
        NumType=Mult                 &             &  \checkmark &             &             &             &             &             \\
        NumType=Ord                  &             &  \checkmark &  \checkmark &             &             &  \checkmark &  \checkmark \\
        Number=Dual                  &             &             &             &             &             &             &             \\
        \textbf{Number=Plur}         &  \checkmark &  \checkmark &  \checkmark &  \checkmark &  \checkmark &  \checkmark &  \checkmark \\
        \textbf{Number=Sing}         &  \checkmark &  \checkmark &  \checkmark &  \checkmark &             &  \checkmark &  \checkmark \\
        Number[psor]=Plur            &             &  \checkmark &  \checkmark &             &             &  \checkmark &  \checkmark \\
        Number[psor]=Sing            &             &  \checkmark &  \checkmark &             &             &  \checkmark &  \checkmark \\
        PartForm=Agt                 &             &             &  \checkmark &             &             &             &             \\
        PartForm=Neg                 &             &             &             &             &             &             &             \\
        PartForm=Past                &             &             &  \checkmark &             &             &             &             \\
        PartForm=Pres                &             &             &  \checkmark &             &             &             &             \\
        PartType=Gen                 &  \checkmark &             &             &             &             &             &             \\
        PartType=Inf                 &  \checkmark &             &             &             &             &             &             \\
        PartType=Neg                 &  \checkmark &             &             &             &             &             &             \\
        Person=0                     &             &             &  \checkmark &             &             &             &             \\
        \textbf{Person=1}            &  \checkmark &  \checkmark &  \checkmark &  \checkmark &  \checkmark &  \checkmark &  \checkmark \\
        \textbf{Person=2}            &  \checkmark &  \checkmark &  \checkmark &  \checkmark &  \checkmark &  \checkmark &  \checkmark \\
        \textbf{Person=3}            &  \checkmark &  \checkmark &  \checkmark &  \checkmark &  \checkmark &  \checkmark &  \checkmark \\
        Person[psor]=1               &             &             &  \checkmark &             &             &             &  \checkmark \\
        Person[psor]=2               &             &             &  \checkmark &             &             &             &  \checkmark \\
        Person[psor]=3               &             &             &  \checkmark &             &             &             &  \checkmark \\
        \textbf{Polarity=Neg}        &             &  \checkmark &  \checkmark &  \checkmark &  \checkmark &  \checkmark &  \checkmark \\
        Polarity=Pos                 &             &             &             &  \checkmark &             &             &  \checkmark \\
        \textbf{Polite=Form}         &             &             &             &             &  \checkmark &  \checkmark &  \checkmark \\
        Polite=Infm                  &             &             &             &             &             &             &  \checkmark \\
        Poss=Yes                     &  \checkmark &  \checkmark &             &             &             &  \checkmark &             \\
        Prefix=Yes                   &             &             &             &  \checkmark &             &             &             \\
        PrepCase=Npr                 &             &             &             &             &             &  \checkmark &             \\
      \pagebreak\hiderowcolors
      \multicolumn{8}{l}{\emph{Continuation of Table A1:}} \\ \\
      \toprule
        Feature Labels               &   Afrikaans &    Croatian &     Finnish &      Hebrew &      Korean &     Spanish &     Turkish \\
      \midrule\showrowcolors
        PrepCase=Pre                 &             &             &             &             &             &  \checkmark &             \\
        \textbf{PronType=Art}        &  \checkmark &             &             &  \checkmark &             &  \checkmark &             \\
        \textbf{PronType=Dem}        &  \checkmark &  \checkmark &  \checkmark &  \checkmark &             &  \checkmark &  \checkmark \\
        PronType=Emp                 &             &             &             &  \checkmark &             &             &             \\
        \textbf{PronType=Ind}        &  \checkmark &  \checkmark &  \checkmark &  \checkmark &             &  \checkmark &  \checkmark \\
        \textbf{PronType=Int}        &  \checkmark &  \checkmark &  \checkmark &  \checkmark &  \checkmark &  \checkmark &             \\
        \textbf{PronType=Neg}        &             &  \checkmark &             &             &             &  \checkmark &             \\
        \textbf{PronType=Prs}        &  \checkmark &  \checkmark &  \checkmark &  \checkmark &             &  \checkmark &  \checkmark \\
        \textbf{PronType=Rcp}        &             &             &  \checkmark &             &             &             &             \\
        \textbf{PronType=Rel}        &  \checkmark &  \checkmark &  \checkmark &             &             &  \checkmark &             \\
        \textbf{PronType=Tot}        &             &  \checkmark &             &             &             &  \checkmark &             \\
        \textbf{Reflex=Yes}          &  \checkmark &  \checkmark &  \checkmark &  \checkmark &             &  \checkmark &  \checkmark \\
        Subcat=Intr                  &  \checkmark &             &             &             &             &             &             \\
        Subcat=Prep                  &  \checkmark &             &             &             &             &             &             \\
        Subcat=Tran                  &  \checkmark &             &             &             &             &             &             \\
        \textbf{Tense=Fut}           &             &             &             &  \checkmark &  \checkmark &  \checkmark &  \checkmark \\
        Tense=Imp                    &             &  \checkmark &             &             &             &  \checkmark &             \\
        \textbf{Tense=Past}          &  \checkmark &  \checkmark &  \checkmark &  \checkmark &  \checkmark &  \checkmark &  \checkmark \\
        Tense=Pqp                    &             &             &             &             &             &             &  \checkmark \\
        \textbf{Tense=Pres}          &  \checkmark &  \checkmark &  \checkmark &             &             &  \checkmark &  \checkmark \\
        VerbForm=Conv                &             &  \checkmark &             &             &             &             &  \checkmark \\
        VerbForm=Fin                 &  \checkmark &  \checkmark &  \checkmark &             &  \checkmark &  \checkmark &             \\
        VerbForm=Ger                 &             &             &             &             &  \checkmark &  \checkmark &             \\
        VerbForm=Inf                 &  \checkmark &  \checkmark &  \checkmark &  \checkmark &             &  \checkmark &             \\
        VerbForm=Part                &  \checkmark &  \checkmark &  \checkmark &  \checkmark &             &  \checkmark &  \checkmark \\
        VerbForm=Vnoun               &             &             &             &             &             &             &  \checkmark \\
        VerbType=Aux                 &  \checkmark &             &             &             &             &             &             \\
        VerbType=Cop                 &  \checkmark &             &             &  \checkmark &             &             &             \\
        VerbType=Mod                 &  \checkmark &             &             &  \checkmark &             &             &             \\
        VerbType=Pas                 &  \checkmark &             &             &             &             &             &             \\
        \textbf{Voice=Act}           &             &  \checkmark &  \checkmark &  \checkmark &             &             &             \\
        Voice=Cau                    &             &             &             &             &  \checkmark &             &  \checkmark \\
        Voice=Mid                    &             &             &             &  \checkmark &             &             &             \\
        \textbf{Voice=Pass}          &             &  \checkmark &  \checkmark &  \checkmark &  \checkmark &             &  \checkmark \\
      \bottomrule
    \end{longtable}
    
    \break

    \rowcolors{1}{white}{whitesmoke}
    \begin{longtable}{lC{1.5cm}C{1.5cm}C{1.5cm}C{1.5cm}C{1.5cm}C{1.5cm}}
      \caption*{Table A2: The monolingual and multilingual probes were evaluated on seven ``held-out'' languages.} \\
      \toprule\hiderowcolors
        Feature Labels               &      Arabic &     Chinese &     Marathi &   Slovenian &     Tagalog &      Yoruba \\
      \midrule\showrowcolors
        \textbf{ADJ}                 &  \checkmark &  \checkmark &  \checkmark &  \checkmark &  \checkmark &  \checkmark \\
        \textbf{ADP}                 &  \checkmark &  \checkmark &  \checkmark &  \checkmark &  \checkmark &  \checkmark \\
        \textbf{ADV}                 &  \checkmark &  \checkmark &  \checkmark &  \checkmark &  \checkmark &  \checkmark \\
        \textbf{AUX}                 &  \checkmark &  \checkmark &  \checkmark &  \checkmark &  \checkmark &  \checkmark \\
        \textbf{CCONJ}               &  \checkmark &  \checkmark &  \checkmark &  \checkmark &             &  \checkmark \\
        \textbf{DET}                 &  \checkmark &  \checkmark &  \checkmark &  \checkmark &  \checkmark &  \checkmark \\
        \textbf{NOUN}                &  \checkmark &  \checkmark &  \checkmark &  \checkmark &  \checkmark &  \checkmark \\
        \textbf{NUM}                 &  \checkmark &  \checkmark &  \checkmark &  \checkmark &             &  \checkmark \\
        \textbf{PART}                &  \checkmark &  \checkmark &  \checkmark &  \checkmark &  \checkmark &  \checkmark \\
        \textbf{PRON}                &  \checkmark &  \checkmark &  \checkmark &  \checkmark &  \checkmark &  \checkmark \\
        \textbf{PROPN}               &  \checkmark &  \checkmark &  \checkmark &  \checkmark &  \checkmark &  \checkmark \\
        \textbf{SCONJ}               &  \checkmark &  \checkmark &  \checkmark &  \checkmark &  \checkmark &  \checkmark \\
        \textbf{VERB}                &  \checkmark &  \checkmark &  \checkmark &  \checkmark &  \checkmark &  \checkmark \\
        AdjType=Attr                 &             &             &             &             &             &             \\
        AdjType=Pred                 &             &             &             &             &             &             \\
        AdpType=Post                 &             &             &             &             &             &             \\
        AdpType=Prep                 &             &             &             &             &             &             \\
        AdpType=Preppron             &             &             &             &             &             &             \\
        AdvType=Tim                  &             &             &             &             &             &             \\
        Animacy=Anim                 &             &             &             &             &             &             \\
        Animacy=Inan                 &             &             &             &             &             &             \\
        Aspect=Hab                   &             &             &             &             &             &             \\
        Aspect=Perf                  &             &             &             &             &             &             \\
        Aspect=Prog                  &             &             &             &             &             &             \\
        Aspect=Prosp                 &             &             &             &             &             &             \\
        Aspect=Rapid                 &             &             &             &             &             &             \\
        \textbf{Case=Abe}            &             &             &             &             &             &             \\
        \textbf{Case=Abl}            &             &             &             &             &             &             \\
        \textbf{Case=Acc}            &  \checkmark &             &  \checkmark &  \checkmark &             &  \checkmark \\
        \textbf{Case=Ade}            &             &             &             &             &             &             \\
        Case=Advb                    &             &             &             &             &             &             \\
        \textbf{Case=All}            &             &             &             &             &             &             \\
        \textbf{Case=Com}            &             &             &             &             &             &             \\
        Case=Comp                    &             &             &             &             &             &             \\
        \textbf{Case=Dat}            &             &             &  \checkmark &  \checkmark &  \checkmark &             \\
        \textbf{Case=Ela}            &             &             &             &             &             &             \\
        \textbf{Case=Equ}            &             &             &             &             &             &             \\
        \textbf{Case=Ess}            &             &             &             &             &             &             \\
        \textbf{Case=Gen}            &  \checkmark &  \checkmark &             &  \checkmark &             &  \checkmark \\
        \textbf{Case=Ill}            &             &             &             &             &             &             \\
        \textbf{Case=Ine}            &             &             &             &             &             &             \\
        \textbf{Case=Ins}            &             &             &  \checkmark &  \checkmark &             &             \\
        \textbf{Case=Loc}            &             &             &  \checkmark &  \checkmark &  \checkmark &             \\
        \textbf{Case=Nom}            &  \checkmark &             &  \checkmark &  \checkmark &             &  \checkmark \\
        \textbf{Case=Par}            &             &             &             &             &             &             \\
        \textbf{Case=Tem}            &             &             &             &             &             &             \\
        \textbf{Case=Tra}            &             &             &             &             &             &             \\
        \textbf{Case=Voc}            &             &             &  \checkmark &             &             &             \\
      \pagebreak\hiderowcolors
      \multicolumn{7}{l}{\emph{Continuation of Table A2:}} \\ \\
      \toprule
        Feature Labels               &      Arabic &     Chinese &     Marathi &   Slovenian &     Tagalog &      Yoruba \\
      \midrule\showrowcolors
        Clitic=Han                   &             &             &             &             &             &             \\
        Clitic=Ka                    &             &             &             &             &             &             \\
        Clitic=Kaan                  &             &             &             &             &             &             \\
        Clitic=Kin                   &             &             &             &             &             &             \\
        Clitic=Ko                    &             &             &             &             &             &             \\
        Clitic=Pa                    &             &             &             &             &             &             \\
        Clitic=S                     &             &             &             &             &             &             \\
        Connegative=Yes              &             &             &             &             &             &             \\
        Definite=Cons                &             &             &             &             &             &             \\
        Definite=Def                 &             &             &             &             &             &             \\
        Definite=Ind                 &             &             &             &             &             &             \\
        Degree=Abs                   &             &             &             &             &             &             \\
        Degree=Cmp                   &             &             &             &             &             &             \\
        Degree=Dim                   &             &             &             &             &             &             \\
        Degree=Pos                   &             &             &             &             &             &             \\
        Degree=Sup                   &             &             &             &             &             &             \\
        Derivation=Inen              &             &             &             &             &             &             \\
        Derivation=Ja                &             &             &             &             &             &             \\
        Derivation=Lainen            &             &             &             &             &             &             \\
        Derivation=Llinen            &             &             &             &             &             &             \\
        Derivation=Minen             &             &             &             &             &             &             \\
        Derivation=Sti               &             &             &             &             &             &             \\
        Derivation=Tar               &             &             &             &             &             &             \\
        Derivation=Ton               &             &             &             &             &             &             \\
        Derivation=Ttain             &             &             &             &             &             &             \\
        Derivation=U                 &             &             &             &             &             &             \\
        Derivation=Vs                &             &             &             &             &             &             \\
        Echo=Rdp                     &             &             &             &             &             &             \\
        Evident=Nfh                  &             &             &             &             &             &             \\
        Form=Adn                     &             &             &             &             &             &             \\
        Form=Aux                     &             &             &             &             &             &             \\
        Form=Compl                   &             &             &             &             &             &             \\
        \textbf{Gender=Fem}          &  \checkmark &             &  \checkmark &  \checkmark &  \checkmark &             \\
        \textbf{Gender=Masc}         &  \checkmark &             &  \checkmark &  \checkmark &  \checkmark &             \\
        \textbf{Gender=Neut}         &             &             &  \checkmark &  \checkmark &             &             \\
        Gender[psor]=Fem             &             &             &             &             &             &             \\
        Gender[psor]=Masc            &             &             &             &             &             &             \\
        Gender[psor]=Neut            &             &             &             &             &             &             \\
        \textbf{HebBinyan=HIFIL}     &             &             &             &             &             &             \\
        \textbf{HebBinyan=HITPAEL}   &             &             &             &             &             &             \\
        \textbf{HebBinyan=HUFAL}     &             &             &             &             &             &             \\
        \textbf{HebBinyan=NIFAL}     &             &             &             &             &             &             \\
        \textbf{HebBinyan=PAAL}      &             &             &             &             &             &             \\
        \textbf{HebBinyan=PIEL}      &             &             &             &             &             &             \\
        \textbf{HebBinyan=PUAL}      &             &             &             &             &             &             \\
        \textbf{HebExistential=True} &             &             &             &             &             &             \\
        \textbf{InfForm=1}           &             &             &             &             &             &             \\
        \textbf{InfForm=2}           &             &             &             &             &             &             \\
      \pagebreak\hiderowcolors
      \multicolumn{7}{l}{\emph{Continuation of Table A2:}} \\ \\
      \toprule
        Feature Labels               &      Arabic &     Chinese &     Marathi &   Slovenian &     Tagalog &      Yoruba \\
      \midrule\showrowcolors
        \textbf{InfForm=3}           &             &             &             &             &             &             \\
        \textbf{Mood=Cnd}            &             &             &             &  \checkmark &             &             \\
        Mood=Des                     &             &             &             &             &             &             \\
        Mood=Gen                     &             &             &             &             &             &             \\
        \textbf{Mood=Imp}            &             &             &  \checkmark &  \checkmark &             &             \\
        \textbf{Mood=Ind}            &  \checkmark &             &  \checkmark &  \checkmark &  \checkmark &             \\
        Mood=Nec                     &             &             &             &             &             &             \\
        Mood=Opt                     &             &             &             &             &             &             \\
        Mood=Pot                     &             &             &             &             &             &             \\
        Mood=Sub                     &             &             &             &             &             &             \\
        NumType=Card                 &             &             &             &             &             &             \\
        NumType=Dist                 &             &             &             &             &             &             \\
        NumType=Frac                 &             &             &             &             &             &             \\
        NumType=Mult                 &             &             &             &             &             &             \\
        NumType=Ord                  &             &             &             &             &             &             \\
        Number=Dual                  &             &             &             &             &             &             \\
        \textbf{Number=Plur}         &  \checkmark &  \checkmark &  \checkmark &  \checkmark &  \checkmark &  \checkmark \\
        \textbf{Number=Sing}         &  \checkmark &             &  \checkmark &  \checkmark &  \checkmark &  \checkmark \\
        Number[psor]=Plur            &             &             &             &             &             &             \\
        Number[psor]=Sing            &             &             &             &             &             &             \\
        PartForm=Agt                 &             &             &             &             &             &             \\
        PartForm=Neg                 &             &             &             &             &             &             \\
        PartForm=Past                &             &             &             &             &             &             \\
        PartForm=Pres                &             &             &             &             &             &             \\
        PartType=Gen                 &             &             &             &             &             &             \\
        PartType=Inf                 &             &             &             &             &             &             \\
        PartType=Neg                 &             &             &             &             &             &             \\
        Person=0                     &             &             &             &             &             &             \\
        \textbf{Person=1}            &  \checkmark &  \checkmark &  \checkmark &  \checkmark &  \checkmark &  \checkmark \\
        \textbf{Person=2}            &  \checkmark &  \checkmark &  \checkmark &  \checkmark &             &  \checkmark \\
        \textbf{Person=3}            &  \checkmark &  \checkmark &  \checkmark &  \checkmark &  \checkmark &  \checkmark \\
        Person[psor]=1               &             &             &             &             &             &             \\
        Person[psor]=2               &             &             &             &             &             &             \\
        Person[psor]=3               &             &             &             &             &             &             \\
        \textbf{Polarity=Neg}        &  \checkmark &  \checkmark &  \checkmark &  \checkmark &  \checkmark &             \\
        Polarity=Pos                 &             &             &             &             &             &             \\
        \textbf{Polite=Form}         &             &             &             &             &             &             \\
        Polite=Infm                  &             &             &             &             &             &             \\
        Poss=Yes                     &             &             &             &             &             &             \\
        Prefix=Yes                   &             &             &             &             &             &             \\
        PrepCase=Npr                 &             &             &             &             &             &             \\
        PrepCase=Pre                 &             &             &             &             &             &             \\
        \textbf{PronType=Art}        &             &             &             &             &             &             \\
        \textbf{PronType=Dem}        &  \checkmark &             &  \checkmark &  \checkmark &  \checkmark &  \checkmark \\
        PronType=Emp                 &             &             &             &             &             &             \\
        \textbf{PronType=Ind}        &             &             &             &  \checkmark &             &  \checkmark \\
        \textbf{PronType=Int}        &             &             &  \checkmark &  \checkmark &             &  \checkmark \\
        \textbf{PronType=Neg}        &             &             &             &  \checkmark &             &             \\
      \pagebreak\hiderowcolors
      \multicolumn{7}{l}{\emph{Continuation of Table A2:}} \\ \\
      \toprule
        Feature Labels               &      Arabic &     Chinese &     Marathi &   Slovenian &     Tagalog &      Yoruba \\
      \midrule\showrowcolors
        \textbf{PronType=Prs}        &  \checkmark &             &  \checkmark &  \checkmark &  \checkmark &  \checkmark \\
        \textbf{PronType=Rcp}        &             &             &             &             &             &             \\
        \textbf{PronType=Rel}        &  \checkmark &             &  \checkmark &  \checkmark &             &  \checkmark \\
        \textbf{PronType=Tot}        &             &             &             &  \checkmark &             &             \\
        \textbf{Reflex=Yes}          &             &             &             &             &             &             \\
        Subcat=Intr                  &             &             &             &             &             &             \\
        Subcat=Prep                  &             &             &             &             &             &             \\
        Subcat=Tran                  &             &             &             &             &             &             \\
        \textbf{Tense=Fut}           &             &             &  \checkmark &  \checkmark &             &             \\
        Tense=Imp                    &             &             &             &             &             &             \\
        \textbf{Tense=Past}          &             &             &  \checkmark &             &             &             \\
        Tense=Pqp                    &             &             &             &             &             &             \\
        \textbf{Tense=Pres}          &             &             &  \checkmark &  \checkmark &             &             \\
        VerbForm=Conv                &             &             &             &             &             &             \\
        VerbForm=Fin                 &             &             &             &             &             &             \\
        VerbForm=Ger                 &             &             &             &             &             &             \\
        VerbForm=Inf                 &             &             &             &             &             &             \\
        VerbForm=Part                &             &             &             &             &             &             \\
        VerbForm=Vnoun               &             &             &             &             &             &             \\
        VerbType=Aux                 &             &             &             &             &             &             \\
        VerbType=Cop                 &             &             &             &             &             &             \\
        VerbType=Mod                 &             &             &             &             &             &             \\
        VerbType=Pas                 &             &             &             &             &             &             \\
        \textbf{Voice=Act}           &  \checkmark &             &             &             &             &             \\
        Voice=Cau                    &             &             &             &             &             &             \\
        Voice=Mid                    &             &             &             &             &             &             \\
        \textbf{Voice=Pass}          &  \checkmark &  \checkmark &             &             &             &             \\
      \bottomrule\hiderowcolors
    \end{longtable}

  \newpage
  \section{Multilayer perceptrons} \label{app:mlp}  % B

    In a small post-hoc analysis with \bert-6, we trained multilayer perceptrons with a single hidden layer (MLP-1s) to perform the multilingual morphosyntactic tagging task. As we increased the dimensionality of the hidden layer, the micro-averaged \fone\ performance would approach that of the monolingual probes, but with comparable or worse selectivity. In contrast, the linear multilingual probes consistently exhibited the best selectivity. Tables B1 and B2 convey these results, given  embeddings of the test corpora. \\

    \begin{table*}[h]   % mlp-f1
      \caption*{Table B1: Micro-averaged \fone\ scores from the linear monolingual and multilingual probes (\emph{Mono.} \& \emph{Multi.}) and the multilingual MLP-1 probes with $h=\{16,32,64,128\}$ hidden dimensions.}
      \centering
      \begin{tabular}{lR{1.5cm}R{1.5cm}R{1.5cm}R{1.5cm}R{1.7cm}R{1.5cm}}
        \toprule
                  &  \emph{Mono.} &  \emph{Multi.} &    $h=16$ &    $h=32$ &    $h=64$ &    $h=128$ \\
        \midrule
        Afrikaans &   \textbf{0.95} &    0.91 &      0.89 &      0.91 &      0.93 &       0.94 \\
        Croatian  &   \textbf{0.92} &    0.87 &      0.83 &      0.88 &      0.90 &       0.91 \\
        Finnish   &   \textbf{0.87} &    0.83 &      0.77 &      0.83 &      0.85 &       \textbf{0.87} \\
        Hebrew    &   \textbf{0.87} &    0.84 &      0.81 &      0.84 &      0.86 &       \textbf{0.87} \\
        Spanish   &   \textbf{0.97} &    0.93 &      0.91 &      0.94 &      0.95 &       0.96 \\
        Turkish   &   \textbf{0.83} &    0.76 &      0.71 &      0.77 &      0.80 &       0.82 \\
        \bottomrule
      \end{tabular}
    \end{table*}

    \begin{table*}[h]   % mlp-sel
      \caption*{Table B2: Selectivity scores from the linear monolingual and  multilingual probes (\emph{Mono.} \& \emph{Multi.}) and the multilingual MLP-1 probes with $h=\{16,32,64,128\}$ hidden dimensions.}
      \centering
      \begin{tabular}{lR{1.5cm}R{1.5cm}R{1.5cm}R{1.5cm}R{1.7cm}R{1.5cm}}
        \toprule
                  &  \emph{Mono.} &  \emph{Multi.} &    $h=16$ &    $h=32$ &    $h=64$ &    $h=128$ \\
        \midrule
        Afrikaans &   0.29 &    \textbf{0.50} &      0.37 &      0.29 &      0.27 &       0.27 \\
        Croatian  &   0.42 &    \textbf{0.58} &      0.42 &      0.39 &      0.39 &       0.39 \\
        Finnish   &   0.46 &    \textbf{0.60} &      0.51 &      0.50 &      0.50 &       0.50 \\
        Hebrew    &   0.49 &    \textbf{0.58} &      0.52 &      0.50 &      0.49 &       0.48 \\
        Spanish   &   0.35 &    \textbf{0.50} &      0.35 &      0.31 &      0.30 &       0.30 \\
        Turkish   &   0.46 &    \textbf{0.47} &      0.39 &      0.38 &      0.39 &       0.40 \\
        \bottomrule
      \end{tabular}
    \end{table*}

  \newpage
  \section{Out-of-vocabulary performance} \label{app:oov}  % C

    For the monolingual and multilingual probes, we micro-averaged separate \fone\ scores for the words that were seen during training (in-vocabulary; IV) and those that weren't (out-of-vocabulary; OOV). We then subtracted the OOV scores from the IV scores to quantify how well the probes generalized to unseen words (Table C1). \\

    \begin{figure*}[h]  % iv - oov
      \caption*{Table C1: Generalizability of the monolingual and multilingual probes. The $x$-axes indicate the \bert\ layer. Negative IV-OOV scores indicate instances where the probes performed \emph{better} on OOV tokens than on IV tokens.} 
      \includegraphics[width=\textwidth]{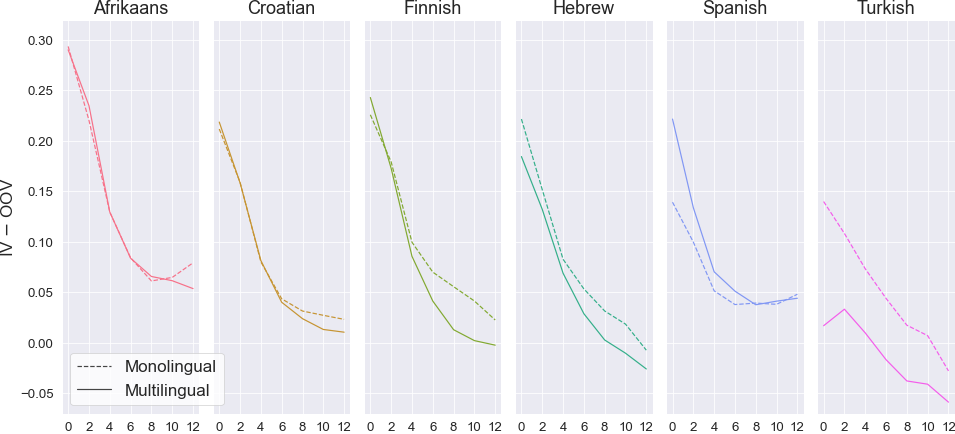}
    \end{figure*}

  \newpage
  \section{Monolingual \& multilingual performance} \label{app:perf}  % D

    \subsection{Afrikaans \fone} \label{app:perf-af}

      \begin{figure}[!h]
        \includegraphics[width=\textwidth]{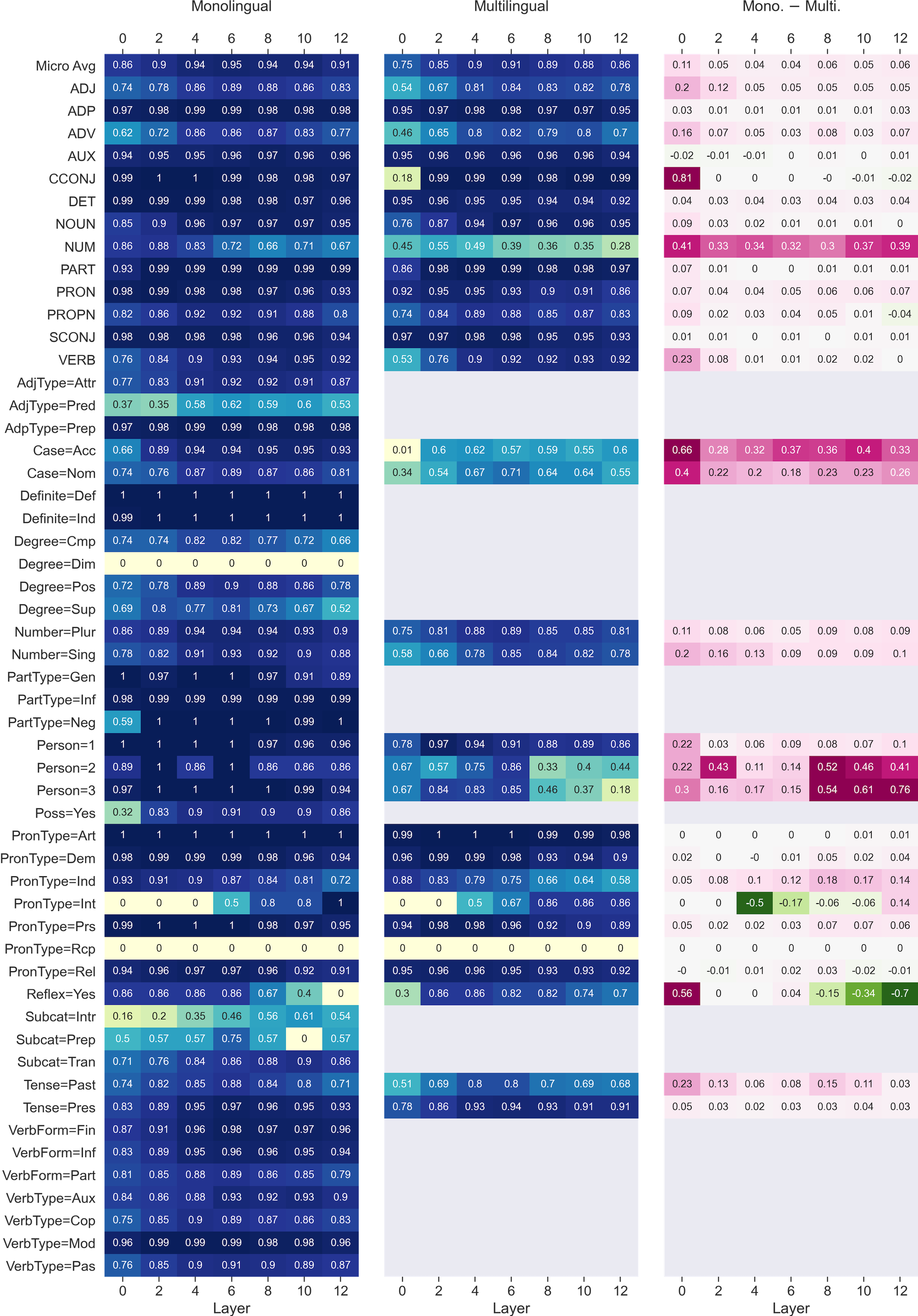}
      \end{figure}
      \vfill\break

    \subsection{Croatian \fone} \label{app:perf-hr}

      \begin{figure}[!h]
        \includegraphics[width=\textwidth]{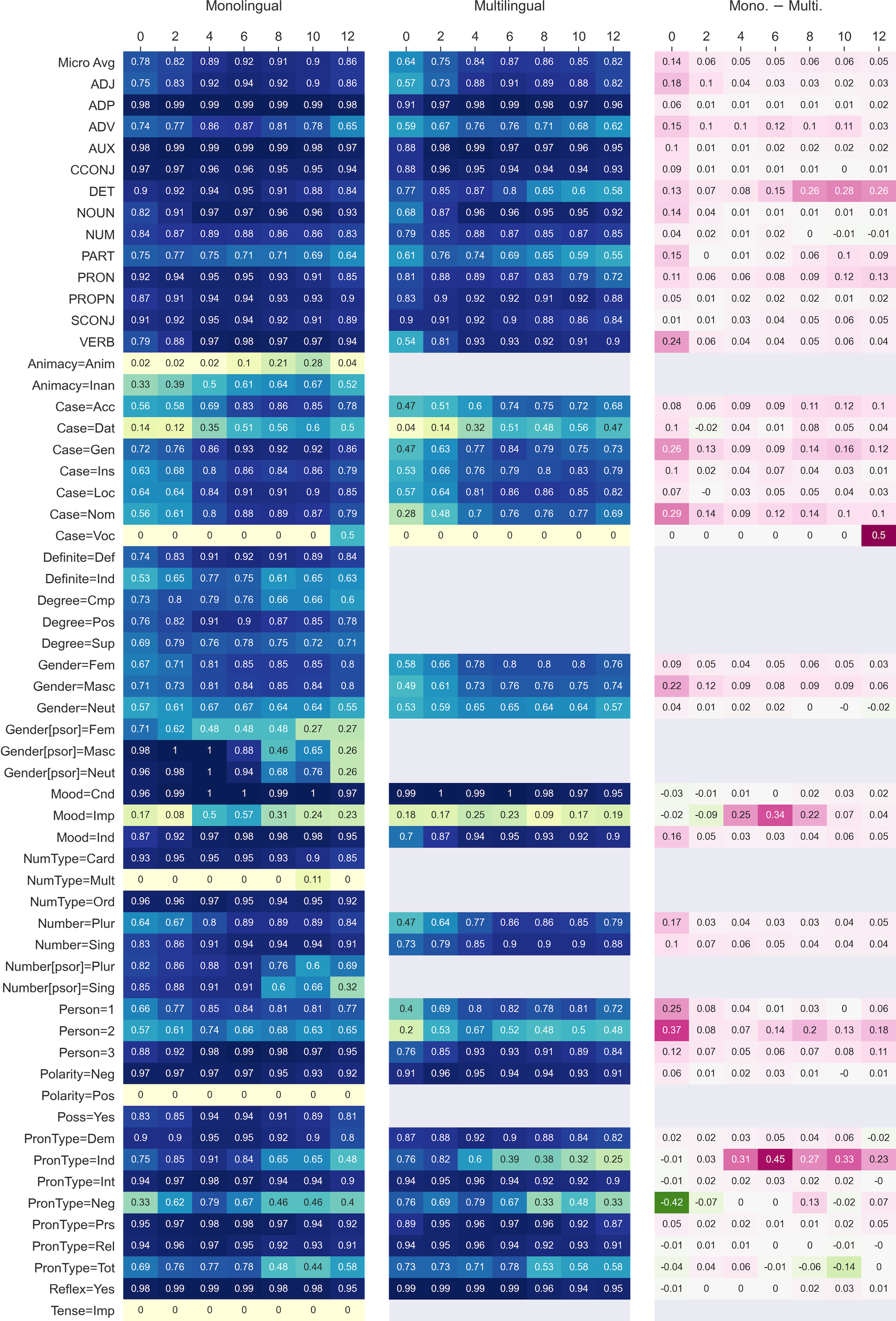}
      \end{figure}
      \begin{figure}[!h]
        \includegraphics[width=\textwidth]{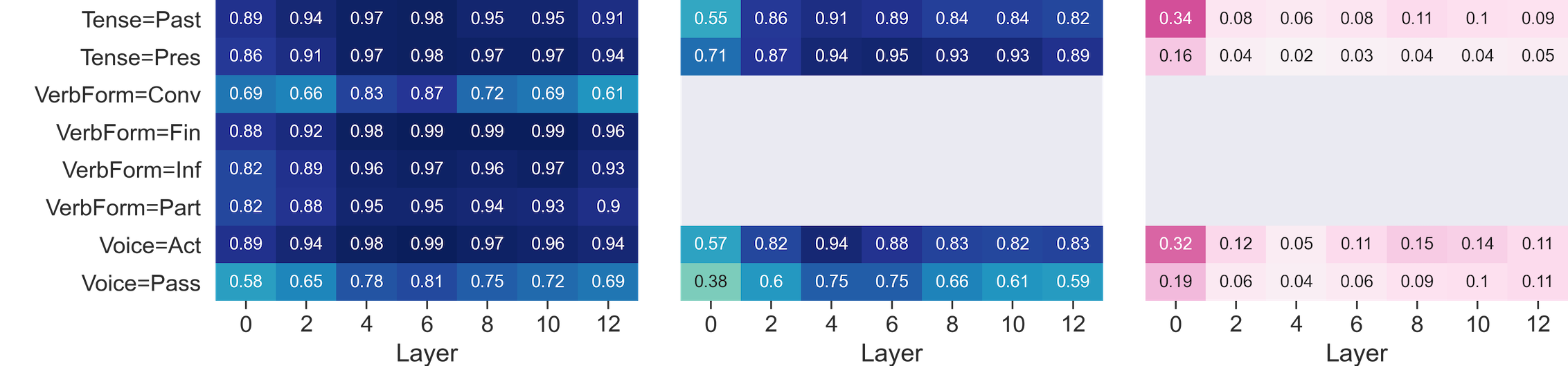}
      \end{figure}
      \vfill\break

    \subsection{Finnish \fone} \label{app:perf-fi}

      \begin{figure}[!h]
        \includegraphics[width=\textwidth]{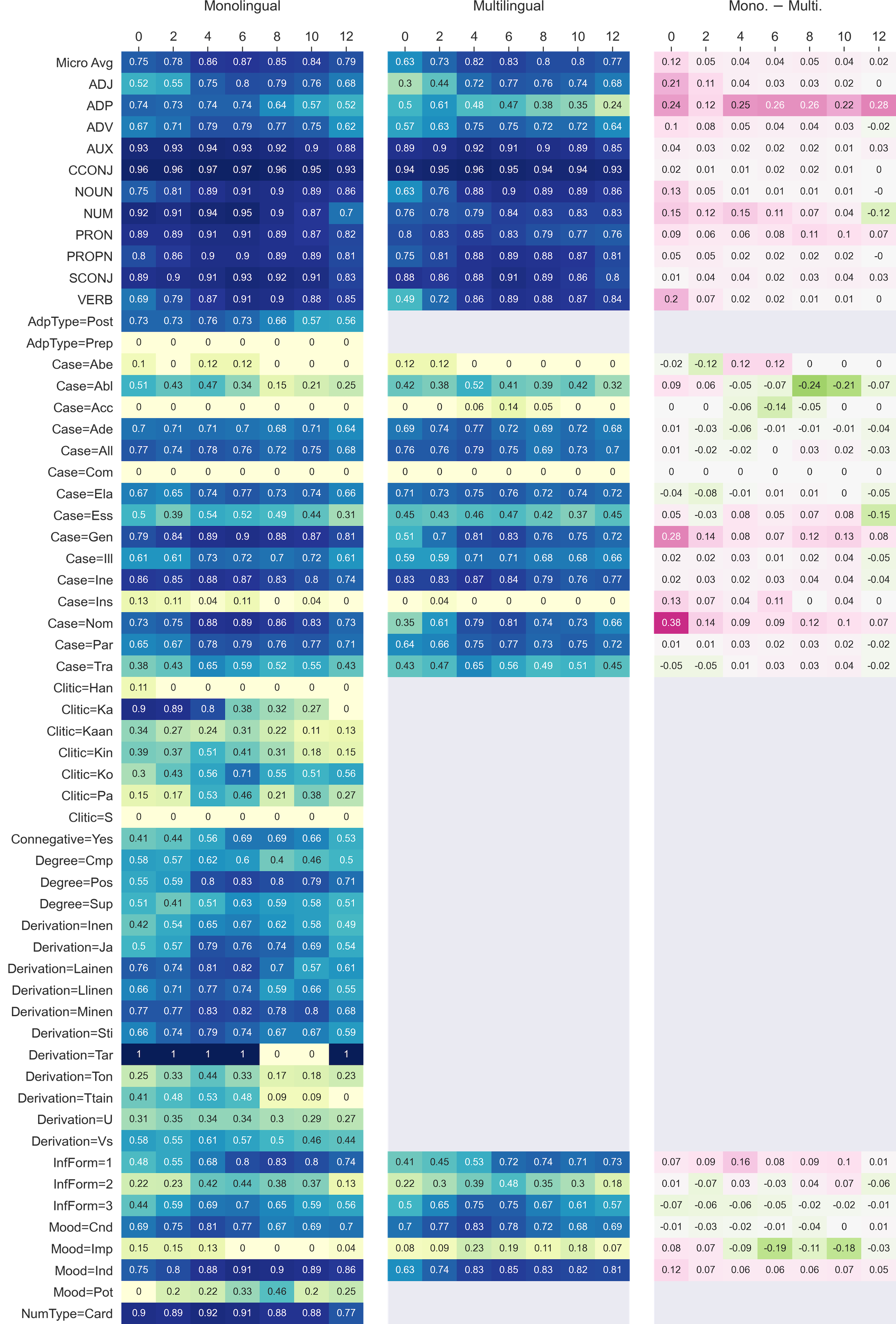}
      \end{figure}
      \begin{figure}[!h]
        \includegraphics[width=\textwidth]{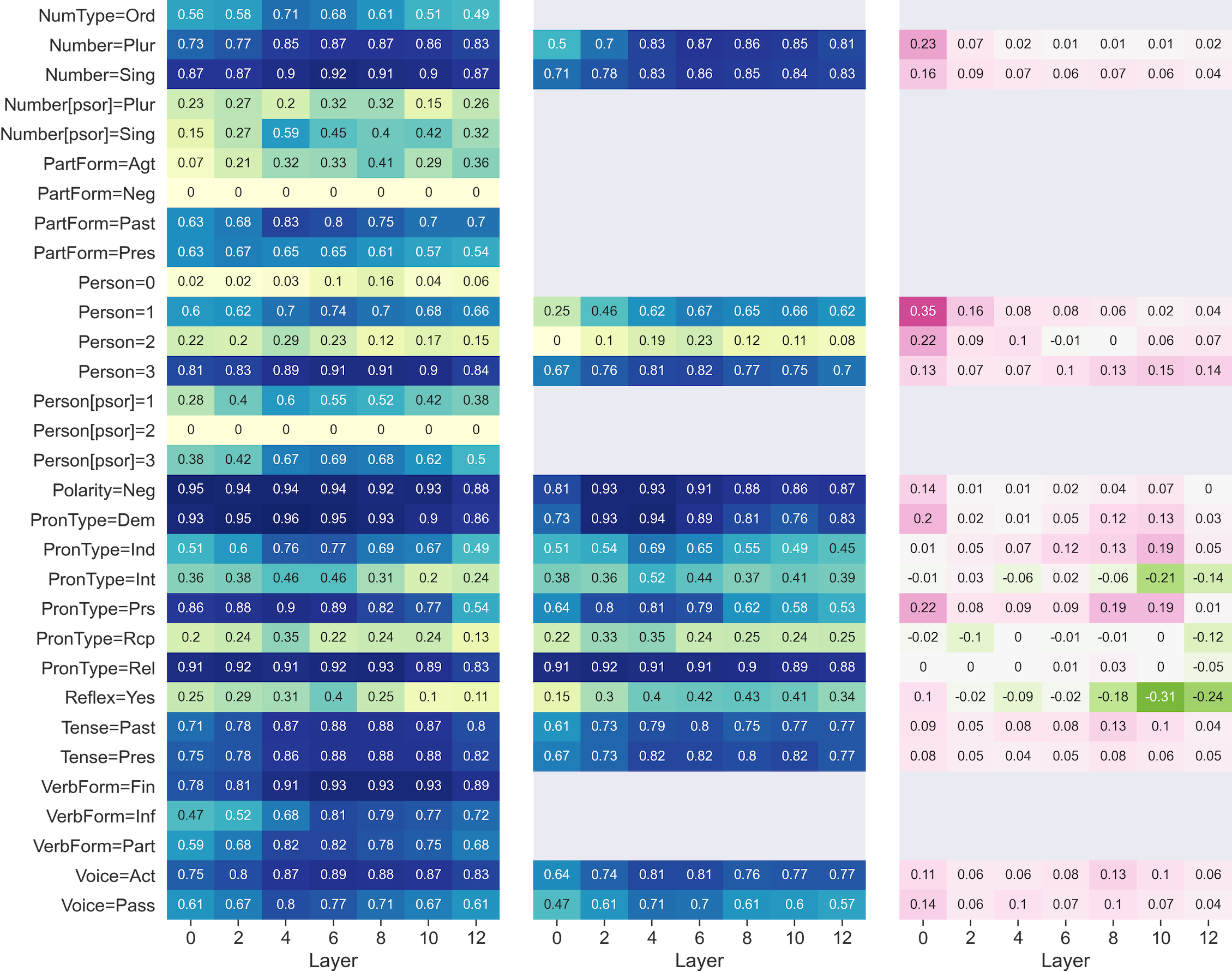}
      \end{figure}
      \vfill\break

    \subsection{Hebrew \fone} \label{app:perf-he}

      \begin{figure}[!h]
        \includegraphics[width=\textwidth]{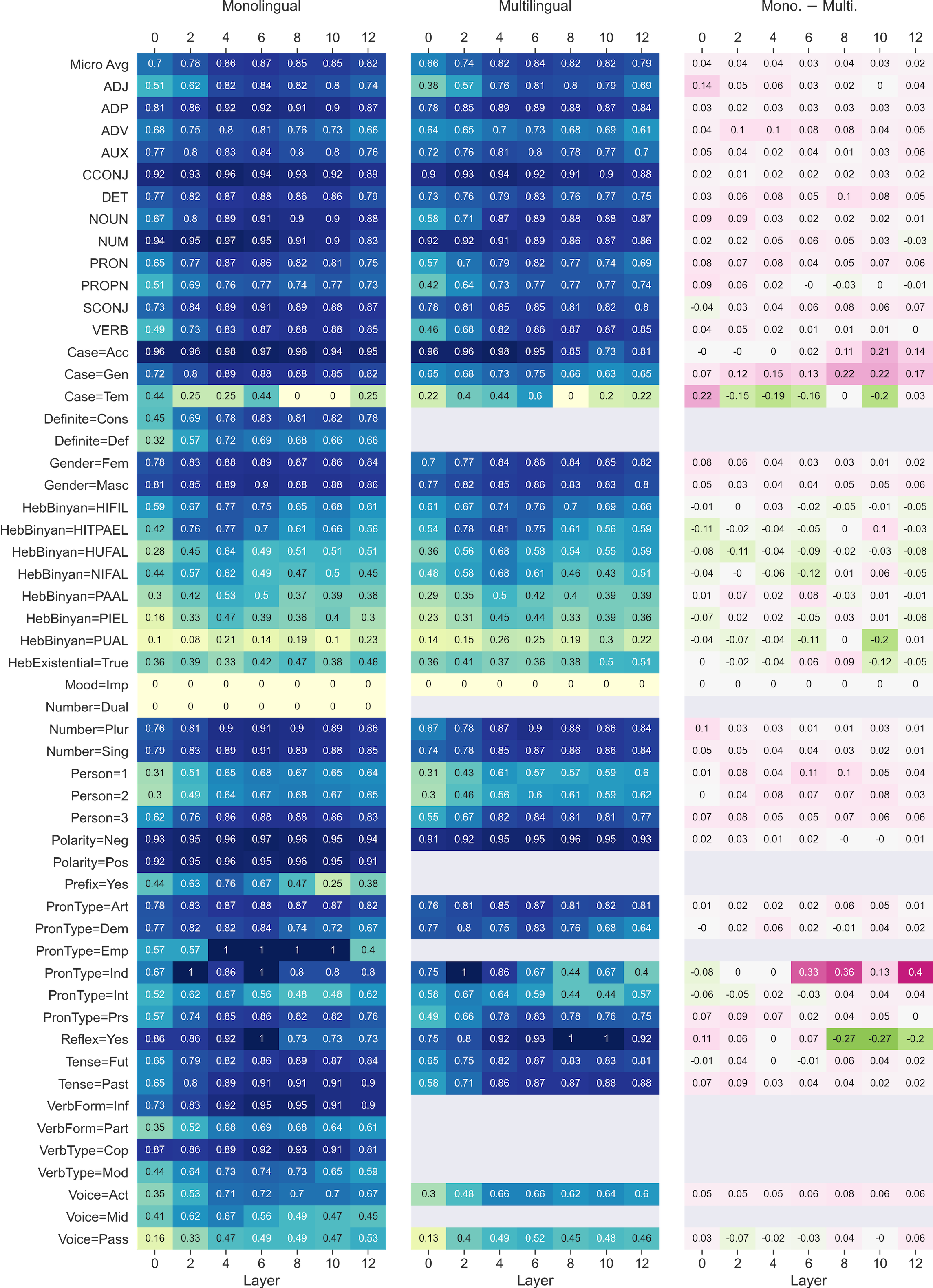}
      \end{figure}
      \vfill\break

    \subsection{Korean \fone} \label{app:perf-ko}

      Korean was not included in the multilingual probes, due to the lack of documentation on the construction of the Korean PUD corpus.\\

      \begin{figure}[!h]
        \includegraphics[width=\textwidth]{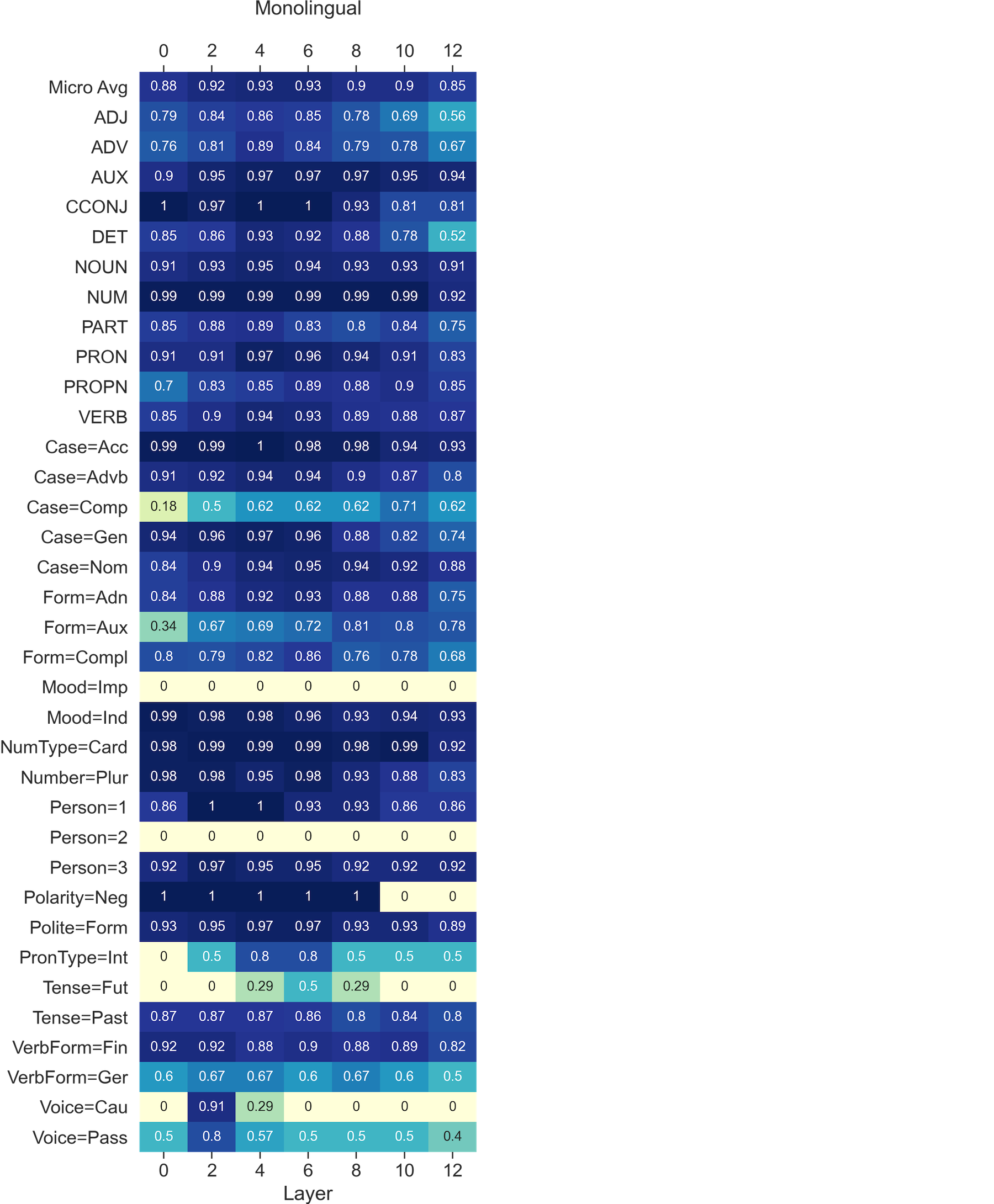}
      \end{figure}
      \vfill\break

    \subsection{Spanish \fone} \label{app:perf-es}

      \begin{figure}[!h]
        \includegraphics[width=\textwidth]{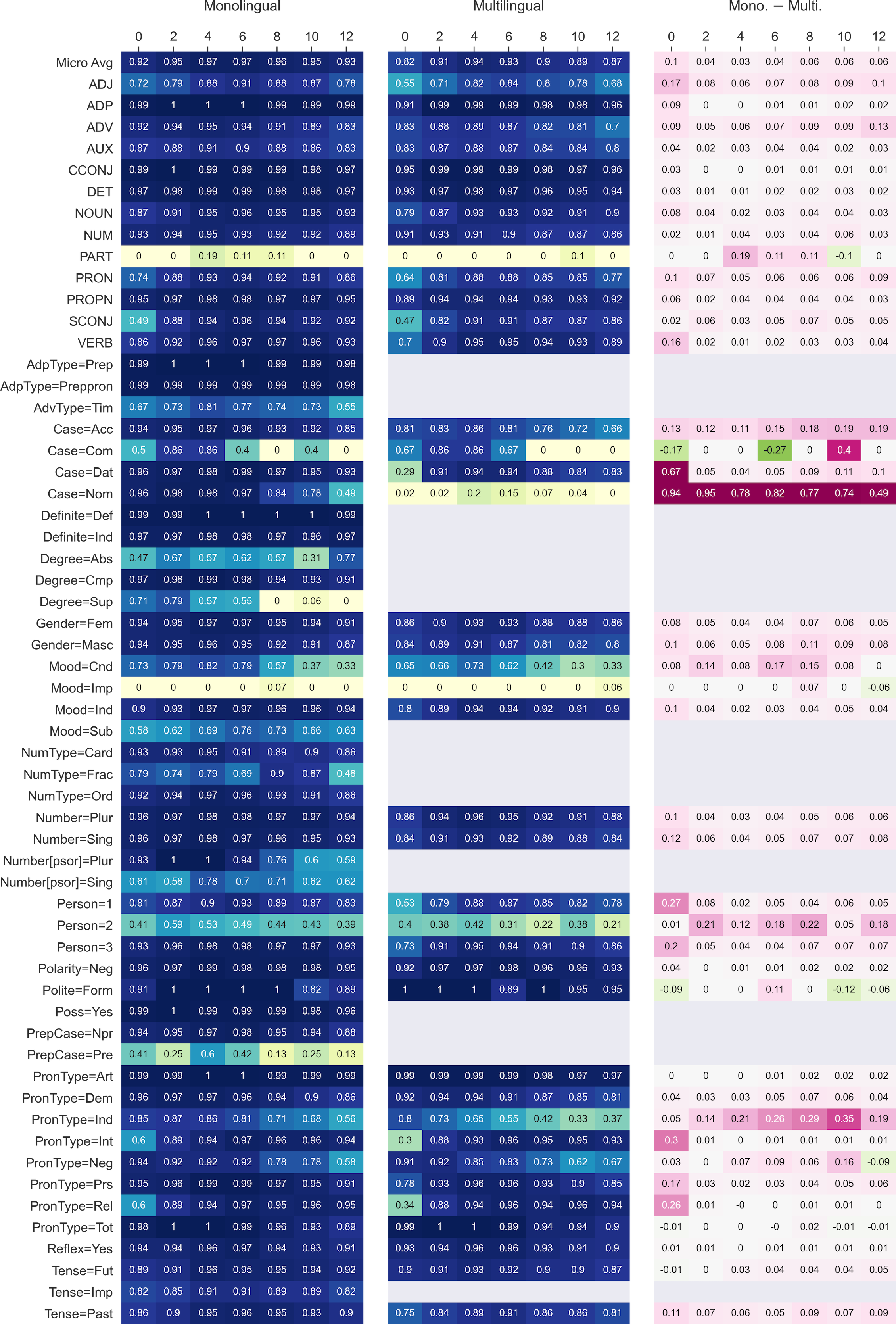}
      \end{figure}
      \begin{figure}[!h]
        \includegraphics[width=\textwidth]{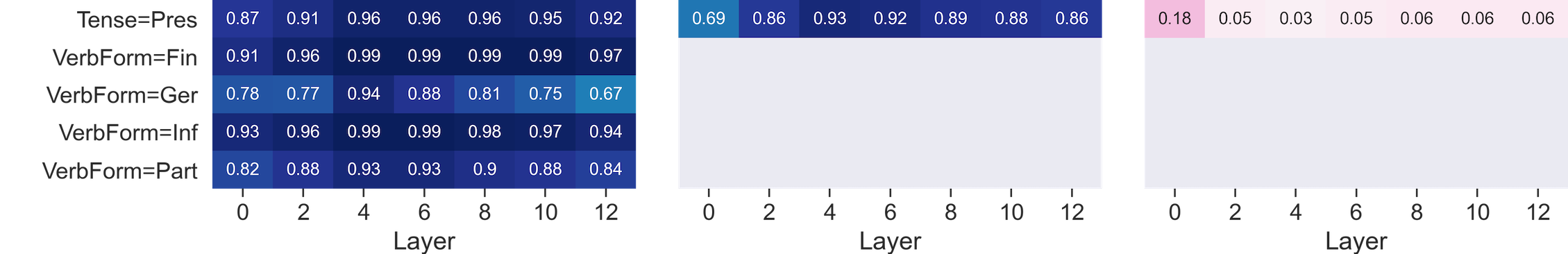}
      \end{figure}
      \vfill\break

    \subsection{Turkish \fone} \label{app:perf-tr}

      \begin{figure}[!h]
        \centering
        \includegraphics[width=\textwidth]{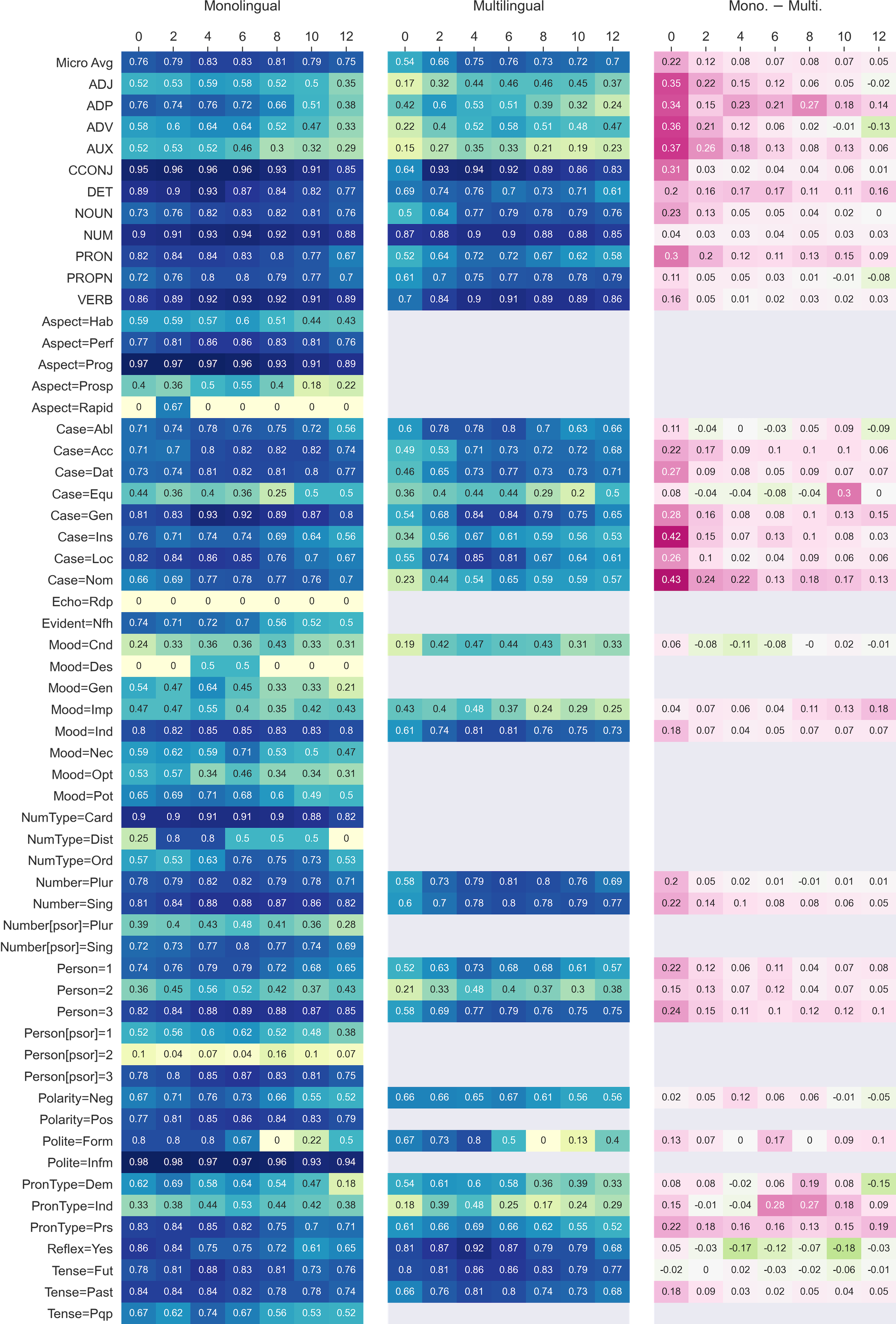}
      \end{figure}
      \begin{figure}[!h]
        \centering
        \includegraphics[width=\textwidth]{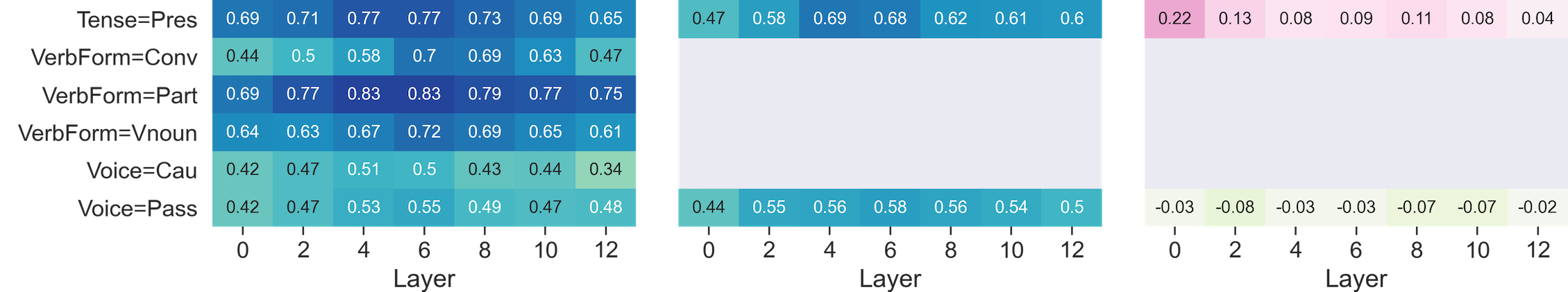}
      \end{figure}
      \vfill

  \newpage
  \section{Crosslingual performance} \label{app:cross}  % E

    The following plots show all of the feature-level \fone\ results from evaluating the monolingual and multilingual \bert-6 probes on the ``held-out'' languages. The $x$-axes indicate the held-out language (Ar=Arabic, Zh=Chinese, Mr=Marathi, Sl=Slovenian, Tl=Tagalog, and Yo=Yoruba) and the $y$-axes indicate the probe (Mu=Multilingual, Af=Afrikaans, Hr=Croatian, Fi=Finnish, He=Hebrew, Ko=Korean, Es=Spanish, and Tr=Turkish). Grayed-out regions indicate where the feature is not applicable to the language or annotated in the language's corpus. \\
  
    \begin{figure}[!h]
      \includegraphics[width=\textwidth]{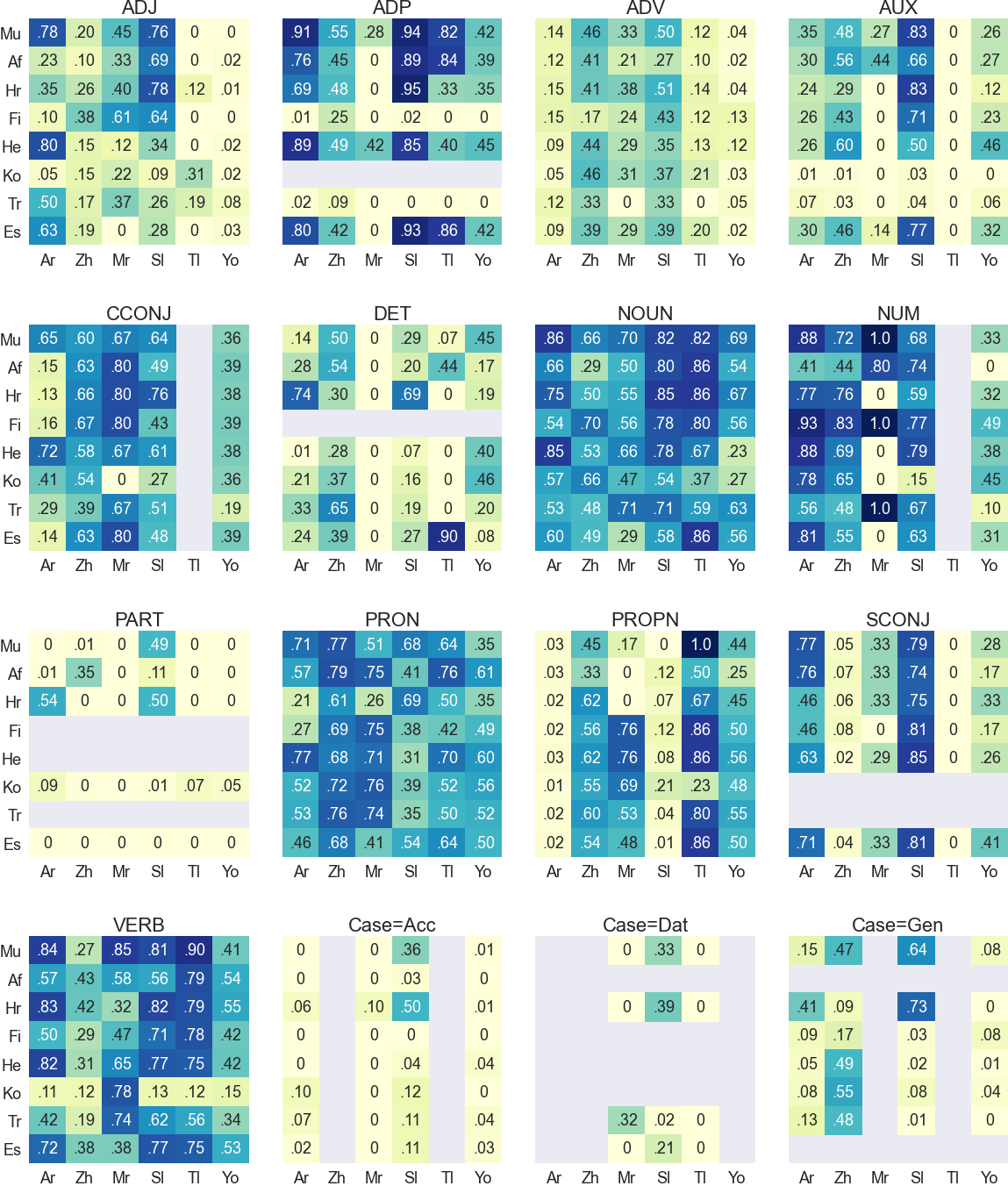}
    \end{figure}
    \break
    \begin{figure}[!ht]
      \includegraphics[width=\textwidth]{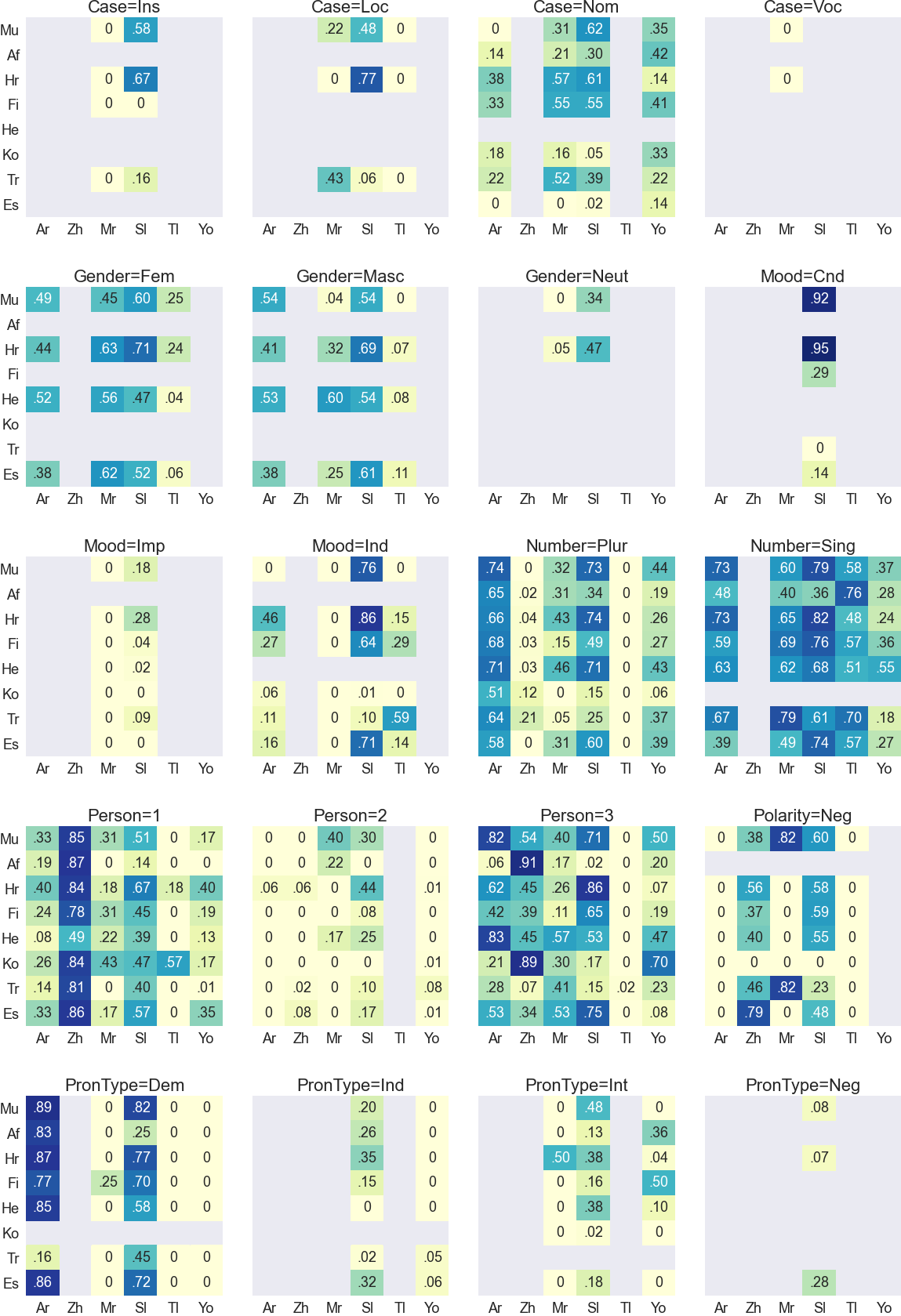}
    \end{figure}
    \break
    \begin{figure}[!ht]
      \includegraphics[width=\textwidth]{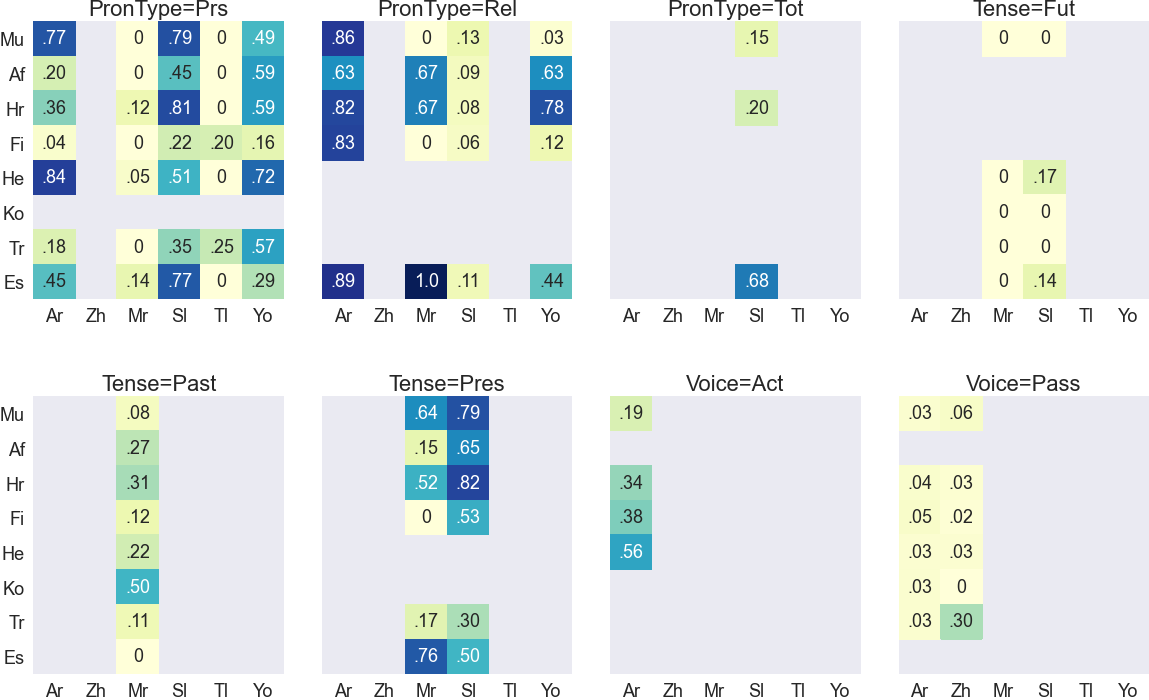}
    \end{figure}
    \break

  \end{appendices}

\end{document}